\documentclass[preprint,12pt]{elsarticle}

\usepackage{amssymb}

\usepackage{diagbox}
\usepackage{amsmath}
\usepackage{amsfonts}
\usepackage{color}
\usepackage{algorithm}
\usepackage[noend]{algpseudocode}
\usepackage{caption}
\usepackage{subcaption}

\journal{Neurocomputing}

\begin{document}

\begin{frontmatter}



\title{UA-PDFL: A Personalized Approach for Decentralized Federated Learning}

\author[jnu,key_lab]{Hangyu Zhu\fnref{eq}}
\author[jnu,key_lab]{Yuxiang Fan\fnref{eq}}
\author[jnu,key_lab]{Zhenping Xie\corref{cor}}

\affiliation[jnu]{organization={School of Artificial Intelligence and Computer Science, Jiangnan University},
            addressline={No.1800 Lihu Road}, 
            city={Wuxi},
            postcode={214122}, 
            state={Jiangsu},
            country={China}}
\affiliation[key_lab]{organization={Jiangsu Key University Laboratory of Software and Media Technology under Human-Computer Cooperation (Jiangnan University)},
            addressline={No.1800 Lihu Road}, 
            city={Wuxi},
            postcode={214122}, 
            state={Jiangsu},
            country={China}}
\fntext[eq]{Equally contributed}
\cortext[cor]{Corresponding author\ead{xiezp@jiangnan.edu.cn}}

\begin{abstract}
Federated learning (FL) is a privacy preserving machine learning paradigm designed to collaboratively learn a global model without data leakage. Specifically, in a typical FL system, the central server solely functions as an coordinator to iteratively aggregate the collected local models trained by each client, potentially introducing single-point transmission bottleneck and security threats. To mitigate this issue, decentralized federated learning (DFL) has been proposed, where all participating clients engage in peer-to-peer communication without a central server. Nonetheless, DFL still suffers from training degradation as FL does due to the non-independent and identically distributed (non-IID) nature of client data. And incorporating personalization layers into DFL may be the most effective solutions to alleviate the side effects caused by non-IID data. Therefore, in this paper, we propose a novel unit representation aided personalized decentralized federated learning framework, named UA-PDFL, to deal with the non-IID challenge in DFL. By adaptively adjusting the level of personalization layers through the guidance of the unit representation, UA-PDFL is able to address the varying degrees of data skew. Based on this scheme, client-wise dropout and layer-wise personalization are proposed to further enhance the learning performance of DFL. Extensive experiments empirically prove the effectiveness of our proposed method.
\end{abstract}

\begin{highlights}
    \item This is the first work that introduces the concept of unit representation, which helps dynamically adjust the number of personalization layers to effectively manage varying degrees of data skew within decentralized federated learning frameworks. Consequently, the proposed method does not rely on public data to precisely capture the nuances of client data distributions, achieving a dual benefit of mitigating communication costs and enhancing system security.
    \item Client-wise dropout mechanism is developed wherein connected clients are dropped out, retaining only one random client when divergence falls below the specified threshold. In this way, the overfitting issue can be effectively mitigated, when client data exhibit relatively independent and identically distributed characteristics.
    \item A layer-wise personalization method is proposed to guide the general feature extractor alongside a personalized classifier, allowing clients to enhance their capability of extracting features while retaining local classification biases.
    \item Empirical experiments are conducted to demonstrate the effectiveness of the proposed UA-PDFL across various dataset and data distributions.
\end{highlights}


\end{frontmatter}


\section{Introduction}
\label{sec:intro}
Federated learning (FL) \citep{fedavg} enables multiple users to jointly learn a shared global model without uploading their private data, significantly alleviating the concerns about privacy leakage. It has been widely used in many real-world applications, such as recommendation systems \citep{HUANG2023118943}, health care \citep{SHEN2023119261, ZHANG2023126791}, Internet-of-Things (IoTs) \citep{9475501,FICCO2024102189}, and so on. However, most existing FL frameworks contain a central server for model aggregation, which is not suitable for scenarios requiring more flexible connectivity, such as unmanned aerial vehicle (UAV) network \citep{9687521}, mobile Ad hoc networks and smart manufacturing \citep{LIU2024}. In addition, the server-client structure of FL is more susceptible to single-point transmission bottlenecks and security threats \citep{9997105,10251949}.

\begin{figure}[ht]
    \centering
    \includegraphics[width=0.92\textwidth]{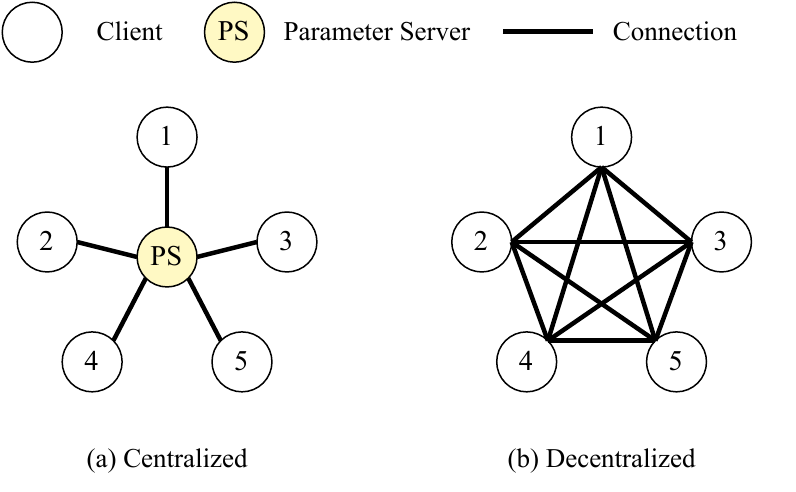}
\caption{A simple example of two different types of FL frameworks. (a) Centralized FL framework containing a parameter server; (b) Decentralized FL framework without a parameter server.}
\label{fig:dfl_comm}
\end{figure}

Therefore, decentralized federated learning (DFL) \citep{roy2019braintorrent,lalitha2018fully} is introduced, where participating clients are peer-to-peer communicated without the requirement of a central server. But eliminating the central server may boost the overall communication costs across the entire learning system. As a simple example depicted in Fig. \ref{fig:dfl_comm}, possible connection paths significantly increases (from 5 paths in Fig. \ref{fig:dfl_comm}(a) to 10 paths in Fig. \ref{fig:dfl_comm}(b)) when clients are peer-to-peer communicated. Hence, minimizing communication overheads while maximizing learning performance become increasingly important, posing a significant challenge for DFL. Some studies have suggested such as employing a proxy model \citep{kalra2023decentralized} and gossip-based pruning \citep{tang2022gossipfl} to reduce the communication costs in DFL. Furthermore, to simultaneously avoid system dominance and address security concerns \citep{zhou2023comavg,gholami2022trusted}, handling data heterogeneity \citep{li2021decentralized, ZHU2021371} and promoting a more sustainable DFL economy system through incentive mechanisms \citep{weng2019deepchain,liu2020fedcoin,NA2024127630} are frequently discussed. However, the inherent non-independent and identically distributed (non-IID) nature of users' training data inevitably impairs the learning performance of DFL \citep{ZHU2021371}. This phenomenon occurs because each client model is prone to converging to different directions, leading to a significant divergence of the aggregated model from the expected ideal model.

To address this issue, maintaining personalized models on each local device during federated training is considered one of the most effective approaches. Among them, some methodologies \citep{rcfl,fedtsdp,comet} utilize public data on server to facilitate either distributed training or client clustering processes. However, the accessibility of public data is not always guaranteed, significantly constraining the scope of application. Other approaches adopt locally computed gradients \citep{cfl,flhc,ditto} or loss function values \citep{ifca} to be clustering metric, obviating the need for public data. Nonetheless, these methods often incur prohibitively high computational costs and fail to adequately capture local representations. In addition, techniques such as sharing only parts of the layers and retaining local personalization layers \citep{fedtsdp,fedper}, performing federated knowledge distillation \citep{wu2022communication,10182241,YU2024127290}, and other strategies \citep{zhang2022personalized,wang2024towards,GUO2023126831} are also regarded as personalized FL (PFL) methods \citep{hanzely2022personalized}. However, these PFL approaches are only applicable to centralized FL frameworks. Therefore, recent work have attempted to integrate the principle of PFL approaches into DFL, utilizing architecture-based methods \citep{dai2022dispfl,ma2022like}, knowledge distillation \citep{10279714} and client selection strategies \citep{ma2022like}. But one key problem is that they are not able to adaptively adjust personalization level to deal with different degrees of client data skew.

In this paper, we propose an \textbf{U}nit representation \textbf{A}ided \textbf{P}ersonalized \textbf{D}ecentralized \textbf{F}ederated \textbf{L}earning (UA-PDFL) framework, which incorporates client-wise dropout and layer-wise personalization to solve the learning degradation challenge caused by non-IID data. The main contributions of this paper are summarized as follows:
\begin{itemize}
    \item This is the first work that introduces the concept of unit representation, which helps dynamically adjust the number of personalization layers to effectively manage varying degrees of data skew within DFL frameworks. Consequently, the proposed method does not rely on public data to precisely capture the nuances of client data distributions, achieving a dual benefit of mitigating communication costs and enhancing system security.
    \item Client-wise dropout mechanism is developed wherein connected clients are dropped out, retaining only one random client when divergence falls below the specified threshold. In this way, the overfitting issue can be effectively mitigated, when client data exhibit relatively IID characteristics.
    \item A layer-wise personalization method is proposed to guide the general feature extractor alongside a personalized classifier, allowing clients to enhance their capability of extracting features while retaining local classification biases.
    \item Empirical experiments are conducted to demonstrate the effectiveness of the proposed UA-PDFL across various dataset and data distributions.
\end{itemize}

\section{Background and Motivation}
In this section, a concise overview of decentralized federated learning is given at first, followed by an introduction to personalized federated learning. Lastly, we reiterate the motivation behind the present work.

\subsection{Decentralized Federated Learning}
\begin{figure}[hp]
    \centering
    \includegraphics[width=0.72\textwidth]{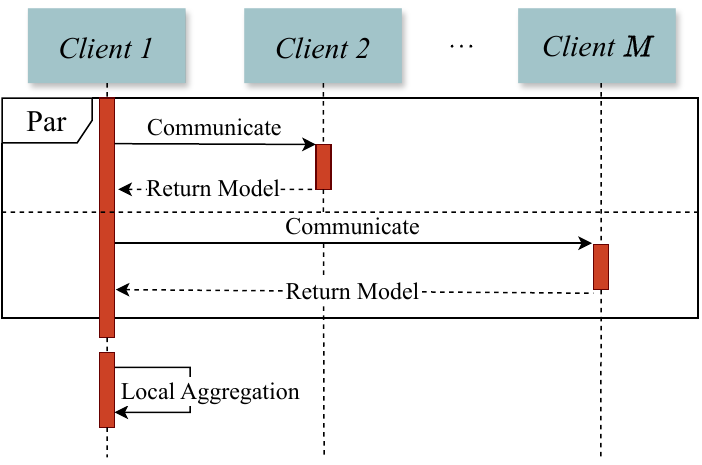}
\caption{Sequence diagram of DFL. Unlike traditional FL, the communication order could be arbitrary.}
\label{fig:dfl_sequence}
\end{figure}

Decentralized federated learning (DFL) \citep{roy2019braintorrent} has been proposed to mitigate issues inherent in the vanilla FL framework, such as centralized server dependencies, transmission bottlenecks, and trustworthiness concerns. As shown in Fig. \ref{fig:dfl_sequence}, $M$ clients conduct local training at first. Subsequently, each client receives model parameters from the other $M-1$ clients for local aggregation, with the communication sequence arbitrarily chosen. One critical challenge in achieving successful DFL lies in addressing communication efficiency. ProxyFL \citep{kalra2023decentralized} presents a classical approach wherein a relatively small model replaces the original one, thereby reducing communication costs. Similarly, GossipFL \citep{tang2022gossipfl} optimizes communication efficiency by selectively canceling non-essential model exchanges based on bandwidth information. Moreover, ensuring system security is paramount. For instance, ComAvg \citep{zhou2023comavg} mitigates potential malicious influences by employing randomized selection of committees, thereby averting undue system domination.

Apart from that, motivating client participation in DFL training is also a critical issue, as ensuring client engagement is imperative for achieving high-quality models. To this end, incentive mechanisms are introduced, wherein participants receive rewards or penalties based on their contributions. DeepChain \citep{weng2019deepchain} exemplifies this approach by implementing a value-driven incentive mechanism leveraging blockchain \citep{zheng2018blockchain} technology to encourage correct participant behavior. Similarly, FedCoin \citep{liu2020fedcoin} facilitates incentive distribution among clients by computing Shapley Values via blockchain. However, these methods, predominantly blockchain-based, may still encounter efficiency concerns.

Moreover, the non-IID problem remains a persistent challenge in DFL. Ma et.al. \citep{ma2022like} proposed a similarity-based approach to guide decentralized connections, motivating clients to connect with those exhibiting similar gradient directions and employing model pruning. Dai et.al. introduced DisPFL \citep{dai2022dispfl}, which adheres the same principle, maintaining a fixed number of active parameters throughout local training and communication. However, these methods do not effectively leverage the information of client data distribution, resulting in significant model aggregation bias throughout the training process.

\subsection{Personalized Federated Learning}
The fundamental concept of PFL-based methods is constructing a personalized model according to the local task to tackle data heterogeneity. In general, there are three major types of PFL methods:

The first approach involves clustering-based methods, which employ a multi-center framework by grouping clients into different clusters. Only clients within the same cluster group participate in model aggregation. CFL \citep{cfl} represents a classical example, performing bi-partitioning to recursively divide participants based on the uploaded model gradient from each client. Similar approaches such as IFCA \citep{ifca} and FL-HC \citep{flhc} utilize computed local loss value as clustering metric for grouping operations \citep{fedtsdp,10081485,long2023multi}. 

The second category encompasses regularization-based methods. For example, Li et al. \cite{ditto} proposed Ditto framework, which involves training a personalized local model subsequent to acquiring a globally generalized model, accomplished by incorporating a regularization term, as shown in Eq.\eqref{eq:ditto}.

\begin{equation}
\label{eq:ditto}
    \min_{w_m}f_m(w_m, w;\mathcal{D}_m)=\mathcal{L}(w_m;\mathcal{D}_m )+\lambda \left \| w_m-w \right \|^{2},
\end{equation}
where $f_m$ represents local training objective on client $m$, $w$ indicates global model parameters and $l$ indicates the local loss function. The hyperparameter $\lambda$ controls the strength of regularization applied during federated training, shrinking the difference between the global model and local models. RCFL \citep{rcfl} and other methods \citep{NEURIPS2020_24389bfe,comet} share similar principles, differing only in the application of various techniques such as utilizing public datasets, among others.

\begin{figure}[ht]
    \centering
    \includegraphics[width=0.92\textwidth]{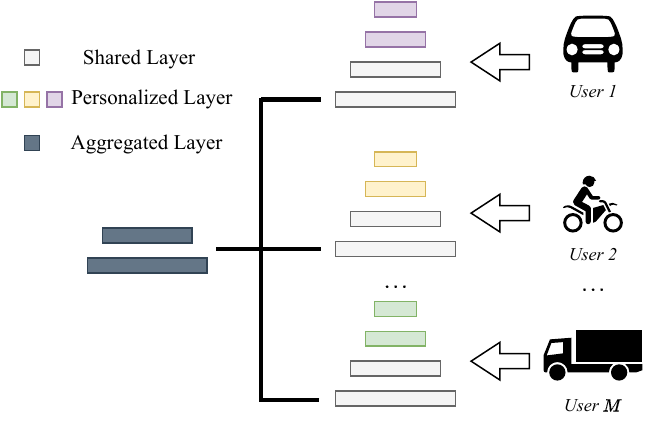}
\caption{An example of FL with personalization layers.}
\label{fig:fedper}
\end{figure}

The third one represents architecture-based methods, allowing each client to possess personalized model structure. Approaches like FedPer \citep{fedper} partition the shared neural network model into personalization layers and base layers (in Fig. \ref{fig:fedper}). Only the base layers need to be uploaded to the server for global model aggregation, significantly preserving the personalized model characteristics. A similar idea is also applied in  LG-FedAvg \citep{liang2020think} and FedAlt \citep{fedalt}, albeit with different personalized positions within the model architecture. In addition, FedMN \citep{10027784} employs a lightweight routing hypernetwork to model the joint distribution on each client and generate personalized selections of module blocks for individual clients.

Indeed, the methods discussed above lack orthogonality and could be further intertwined, rendering the taxonomy non-unique. Nevertheless, in summary, there have been relatively few studies that have integrated personalized learning methods into DFL to deal with the challenge of non-IID data problem.

\section{Proposed Method}
In this section, the technical details of the proposed UA-PDFL would be discussed. The math formulation of personalized decentralized federated learning (PDFL) is given at first. Followed by the description of introduced unit representation, client-wise dropout and layer-wise personalization. Finally, the overall framework of UA-PDFL is illustrated at last.

\subsection{Problem Description}
Assume there are total $M$ clients, each aiming to minimize its local training object $f_{m}(w_{m})$, we can formulate PDFL problem based on literature \citep{fedavg,hanzely2022personalized} as an optimization process described by Eq. \eqref{eq:PDFL}:

\begin{equation}
\label{eq:PDFL}
    \min_{w_{1:M}\in \mathbb{R}^d}{F(w_{1:M})=\sum_{m=1}^{M}f_{m}(w_{m}; \mathcal{D}_m)}=\sum_{m=1}^{M}\mathcal{L}_{\left( \mathcal{X},y \right)\sim \mathcal{D}_{m}}(w_m;\left( \mathcal{X},y \right)),
\end{equation}
where $w_m$ denotes model parameters of client $m$, $F(w_{1:M})$ represents the global loss, which is the summation of local loss $f_{m}(w;\mathcal{D}_m)$, and $\mathbb{R}^d$ denotes the space of model parameters. In summary, the goal of the PDFL is to obtain a set of model weights $w_{1:M}$ to maximize the learning performance on each local client data $\mathcal{D}_{m}$.

Due to the absence of central server, it is necessary for clients to maintain their own communication queue, denoted as $Q_m$, where $|Q_m| \leq M$. And owing to the variations in computational capacities, the transmission of client model occurs asynchronously.

\subsection{Unit Representation}
\label{sec:std_rep}
In the realm of FL, researchers commonly rely on publicly available dataset to discern local data distribution for further client clustering \citep{fedtsdp,comet,flis}. However, in the absence of a central server, maintaining this functionality requires the transmission of public dataset among clients. This practice significantly increases the risk of data leakage due to the involvement of multiple entities in data handling. Furthermore, the reliance on peer-to-peer (P2P) model transmission exacerbates the demands on network bandwidth and availability, thereby posing additional challenges to the scalability and overall efficiency of FL systems. As a result, the straightforward application of such methodologies necessitates careful modification and consideration of these factors.

Due to the fact that parametric models inherently encapsulate the information of the training data \citep{nn_has_memory}, it is unnecessary to rely on public dataset for identifying local data distributions. And we have observed a phenomenon in the sub-experiment wherein neural networks exhibit a similar characteristic of \emph{inertial thinking} as the human brain. That is, for instance, when a neural network model is trained on a specific dataset, the model inferences for any unknown inputs tend to closely resemble the output labels of the training data. For client model $w_{m}$ optimized by the local loss function $\mathcal{L}_{\left( \mathcal{X},y \right)\sim \mathcal{D}_{m}}\left( w_{m};\left( \mathcal{X},y \right) \right)$, the model outputs of unknown sample $\mathcal{X}_{u}$ are likely to approximate $\mathbb{E}_{\left( \mathcal{X},y \right)\sim \mathcal{D}_{m}}[p(y|\mathcal{X})]$. However, it should be noticed that since the inference process of neural networks is not linear, the inference distribution of $\mathcal{X}_{u}$ is not equal to that of the real client data distribution $p(\mathcal{D}_m)$ as shown in Eq. \eqref{eq:local_distribution}:

\begin{equation}
\label{eq:local_distribution}
    p(\mathcal{D}_m) \neq p\left(\sigma\left(f_{m}\left(w_{m};\mathcal{X}_{u} \right) \right) \right),
\end{equation}
where $\sigma$ is the activation function. Apparently, communicating the model inference $p\left(\sigma\left(f_{m}\left(w_{m};\mathcal{X}_{u} \right) \right) \right)$ among participants in DFL would not expose the local label information.
However, the negotiation of $\mathcal{X}_{u}$ remains a significant challenge. Variations in the choice of $\mathcal{X}_{u}$ can result in inconsistent representations of the data distribution. Achieving consensus on $\mathcal{X}_{u}$ necessitates additional communication rounds, and this process further requires the implementation of consensus algorithms to address issues such as network partitioning. Based on the aforementioned analysis, we create an unit tensor $\mathcal{X}_{unit}$ to replace the unknown sample $\mathcal{X}_{u}$ for constructing the representation of data distribution. The unit tensor $\mathcal{X}_{unit}$ is defined as follows:

\begin{equation}
\label{eq:std_input}
    \mathcal{X}_{unit}=(a_{ij})\in\mathbb{R}^{s},
\end{equation}
where constant $a_{ij}$ represents the element at the $i$-th row and $j$-th column of unit tensor $ \mathcal{X}_{unit}$, and $\mathbb{R}^{s}$ is the sample space. The calculation of $a_{ij}$ can be achieved by averaging a subset of data samples or by assigning predetermined constant values, without the necessity of individual data points. Substituting the unknown tensor $\mathcal{X}_{u}$ with a predefined unit tensor $\mathcal{X}_{unit}$ can effectively reduce both the complexity and communication overhead associated with negotiating input tensors among clients. Additionally, as $\mathcal{X}_{unit} \in \mathcal{X}_{u}$, it can inherently and accurately represent the data distribution differences across clients. Then, the unit representation is computed by Eq. \eqref{eq:std_rep}:

\begin{equation}
\label{eq:std_rep}
    I_m=\sigma \left(f_{m}(w_{m};\mathcal{X}_{unit}) \right),
\end{equation}
where $I_m$ stands for the unit representation of client $m$. Since the unit representation implicitly reflects the form of local data distribution, we can adopt Jensen-Shannon divergence \citep{menendez1997jensen} as the divergence metric $\text{Div}(i,j)$ between client $i$ and client $j$:

\begin{equation}
\label{eq:divergence_metric}
    \text{Div}(i,j)=\frac{1}{2}\text{KL}(I_{i}||I_{j})+\frac{1}{2}\text{KL}(I_{j}||I_{i}),
\end{equation}
where $\text{KL}(I_{i}||I_{j})$ is the Kullback–Leibler divergence \citep{kullback1951information} of unit representation of $I_{i}$ and $I_{j}$, with its mathematical formulation defined in Eq. \eqref{eq:kldiv}:

\begin{equation}
\label{eq:kldiv}
    \text{KL}(I_{i}||I_{j})=\sum_{o \in O}I_{i,o}\log{\frac{I_{i,o}}{I_{j,o}}},
\end{equation}
where $o$ is the output index of the model inference and $O$ represents all possible output classes.

\begin{figure}[hbtp]
    \centering
    \includegraphics[width=0.92\textwidth]{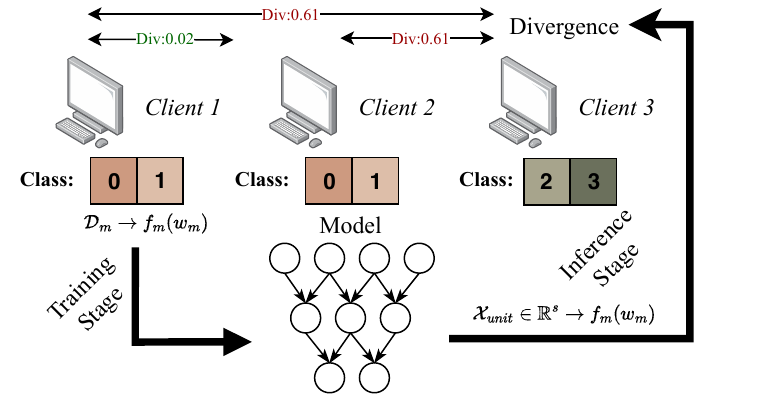}
\caption{The relationship between local data distribution and divergence metric. Firstly, the neural network model is trained by local dataset. And then, the unit representation and divergence metric are obtained by Eq. \eqref{eq:std_rep} and Eq. \eqref{eq:divergence_metric}, respectively.}
\label{fig:standard_rep}
\end{figure}

To verify the effectiveness of divergence computed by our proposed unit representation method, we conducted a toy experiment as shown in Fig. \ref{fig:standard_rep}. Where both client 1 and client 2 hold data samples from class 0 and 1, while client 3 holds data samples from class 2 and 3. And a standard multi-layer perception neural network with shallow layers was selected for client training, with the objective of achieving a local validation accuracy exceeding 90\%. Subsequently, the unit input $\mathcal{X}_{unit}$ with $a_{ij}=1$ was employed to obtain inference responses for calculating the divergence metric among these three clients pairwise. The experimental results indicate that clients with similar data distribution, such as client 1 and 2, exhibit a smaller divergence ($\text{Div}(1, 2)=0.02$) compared to those with larger divergence in data distribution ($\text{Div}(1, 3)=\text{Div}(2, 3)=0.61$). This finding demonstrates a strong correlation between local data distribution and model divergence, and further analysis will be conducted in Section \ref{sec:analysis_std_rep}.

\subsection{Client-wise Dropout}
Dropout is a widely adopted technique for mitigating overfitting in neural networks \citep{dropout}. Previous research work have indicated that dropout is mathematically equivalent to an approximation of a deep Gaussian process \citep{gal2015dropout,pmlr-v48-gal16}. This concept can be extended to DFL framework that each client user can be viewed as a neuron within the model layers. And according to Eq. \eqref{eq:PDFL}, the model aggregation operation can be interpreted as a straightforward linear combination of the feedforward pass. Consequently, Eq. \eqref{eq:PDFL} can be modified to the following Eq. \eqref{eq:dflcw} by incorporating client-wise dropout:

\begin{equation}
\label{eq:dflcw}
    \mathcal{L_{\text{dropout}}}=\sum_{m=1}^{M}f_{m}(\theta_{m}; \mathcal{D}_m)+\lambda \left \| \Theta \right \|_{2}^{2},
\end{equation}
where $\Theta=\left \{ \theta_{1},\theta_{2}, ..., \theta_{M} \right \}$ represents a set of $M$ connection weights. To approximate this model, the predictive distribution of Gaussian process for a new input $\mathbf{x}^{*}$ on $\Theta$ is given by Eq. \eqref{eq:ppred}:

\begin{equation}
\begin{aligned}
\label{eq:ppred}
    p\left ( \mathbf{y}^{*}|\mathbf{x}^{*},\mathbf{X},\mathbf{Y} \right )&=\int p\left ( \mathbf{y}^{*}|\mathbf{x}^{*},\Theta \right )p\left ( \Theta|\mathbf{X},\mathbf{Y} \right )d\Theta \\
    &=\int q(\Theta)\text{log}p\left ( \mathbf{Y}| \mathbf{X},\Theta\right )d\Theta-\text{KL}\left ( q\left ( \Theta \right )||p\left ( \Theta \right ) \right ),
\end{aligned}
\end{equation}
where $\mathbf{X} \in \mathbb{R}^{N \times Q}$ and $\mathbf{Y} \in \mathbb{R}^{N \times 1}$ are $N$ training inputs and labels, respectively. Maximizing the above predictive distribution is equivalent to minimizing $\text{KL}\left ( q\left ( \Theta \right )||p\left ( \Theta \right ) \right )$. According to the theoretical analysis introduced in \citep{gal2015dropout}, Eq. \eqref{eq:ppred} can be approximated into $\mathcal{L}_{\text{GP-MC}}$ by adopting variational inference and Monte Carlo integration:

\begin{equation}
\begin{aligned}
\label{eq:vimc}
    \mathcal{L}_{\text{GP-MC}}&=\sum_{n=1}^{N}\text{log}p\left ( \textbf{y}_{n}| \textbf{x}_{n},\Theta \right )-\text{KL}\left ( q\left ( \Theta \right )||p\left ( \Theta \right ) \right ) \\
    &\propto \frac{1}{N\tau}\sum_{n=1}^{N}-\text{log} p\left ( \mathbf{y}_{n}|\mathbf{x}_{n},\widehat{\theta}_{n} \right )+\frac{P}{2N\tau}\left \| M \right \|_{2}^{2},
\end{aligned}
\end{equation}
where $\widehat{\theta}_{n} \sim q(\Theta)$ is a random connection weight sampled from variational distribution $q(\Theta)$, $M=\Theta \cdot \text{diag}\left ( \left [ \textbf{z}_{j} \right ]_{j=1}^{M} \right )^{-1}$, $\tau$ is the given model precision, and $z_{j} \sim \text{Bernoulli}(P)$. The term $\frac{P}{2N\tau}\left \| M \right \|_{2}^{2}$ can be interpreted as the additional regularization component representing client-wise dropout.Consequently, it is interesting to find that dropout connections in DFL can also be approximated to a Bayesian model.

\begin{figure}[hbtp]
    \centering
    \includegraphics[width=0.92\textwidth]{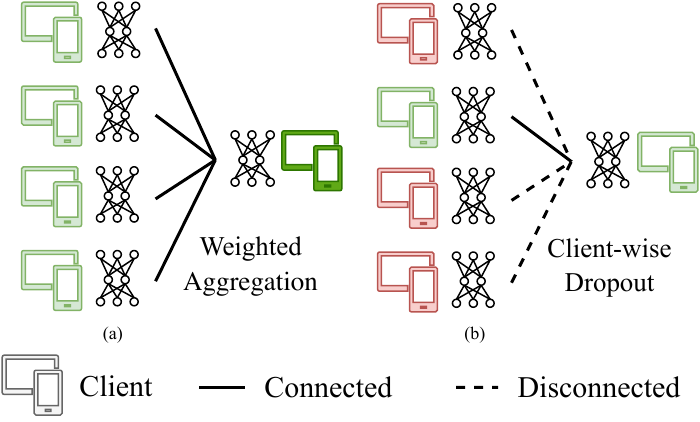}
\caption{A simple example of client-wise dropout. (a) All the clients are connected for model aggregation. (b) Only one client is communicated for model replacement.}
\label{fig:client_wise_dropout}
\end{figure}

Inspired by the principle of client-wise dropout mentioned above, it is advantageous to discard connected clients to facilitate the attainment of a collaborative learning objective. In addition, given that all parametric client models in DFL are typically updated via stochastic gradient descent (SGD) on mini-batch local data, if the entire training data exhibits homogeneity, it is adequate to retain only one connected client instead of applying dropout following the the Bernoulli distribution $z_{j} \sim \text{Bernoulli}(P)$ (Fig. \ref{fig:client_wise_dropout}(b)). In this scenario, DFL process is simplified to a min-batch SGD optimization.

In addition, we utilize an inference threshold $th_{I}$ to regulate the diversity of clients in DFL. And client-wise dropout is executed when $\forall \text{Div}(m,i) < th_{I}, C_i\in Q_m$ to prevent any detrimental effects caused by non-IID data.

\subsection{Layer-wise Personalization}
In transfer learning, to retain prior knowledge while adapting to a new domain, it is common to further decompose a neural network into a feature extractor $g(\cdot)$ and a classifier $h(\cdot)$ \citep{technologies11020040}. The concept of personalization layers in FL \citep{fedper} operates on a similar principle as transfer learning but mantains a fixed number of shared layers. Meanwhile, in scenarios involving multiple participants, the distribution may vary distinctively, rendering it unsuited to share the same number of layers.

\begin{figure}[hbtp]
    \centering
    \includegraphics[width=0.92\textwidth]{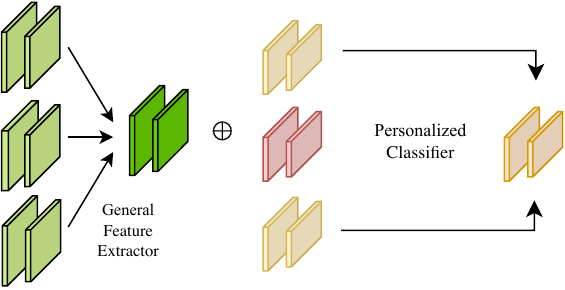}
\caption{An example of layer-wise personalization, where $\bigoplus$ means combining the general feature extractor and personalized classifier into the whole local model.}
\label{fig:layer_wise_agg}
\end{figure}

Motivated by the methods discussed above, we propose a federated layer-wise personalization method, which emphasizes training an universal feature extractor and migrating similar classifiers, thereby adapting the the local personalization capability. This approach is particularly well-suited for DFL settings, where each client retains full control over its private model. The overall mechanism is illustrated in Fig. \ref{fig:layer_wise_agg}, where green planes represent the general feature extractor, and yellow and orange planes denote personalized classifiers. It is important to note that dissimilar classifiers (red plane) do not perform client model aggregation.

To guide the converge direction of the local feature extractor, obtaining the auxiliary representation is imperative. This can be achieved through Eq. \eqref{eq:middle_rep}:

\begin{equation}
\label{eq:middle_rep}
    I^{Aux}_m = g_m(\mathcal{X}_{unit}),
\end{equation}
where $g_{m}$ is the feature extractor on client $m$, and $I^{Aux}_m$ is the corresponding auxiliary representation. For each client, we propose to add a proximal term $\left\|I^{Aux}_m-I^{Aux}_{avg}\right\|_{2}^{2}$ to the original loss function $\mathcal{L}(w_{m};\mathcal{D}_{m})$ to suppress the impact from local updates, where $I^{Aux}_{avg}$ is the average auxiliary representation computed by $C_{i} \in Q_{m}$. To be more specifically, the local loss is then derived in Eq. \eqref{eq:local_object}:

\begin{equation}
\label{eq:local_object}
    f_m(w_m) = \mathcal{L}(w_m;\mathcal{D}_{m}) + \mu\left\|I^{Aux}_m-I^{Aux}_{avg}\right\|_{2}^{2},
\end{equation}
where $\mu$ is the mixture coefficient of the regularization term. In addition, to further enhance the classification performance, we fuses model parameters $w_{h}$ of similar classifiers $h(\cdot)$ when $\text{Div(m,i)}$ is lower than the threshold $th_{I}$. The pseudo code for layer-wise personalization on client $m$ is illustrated in Algorithm \ref{algorithm:layer_agg}.

\begin{algorithm}[H]
\caption{Layer-wise Personalization}\label{algorithm:layer_agg}
\begin{algorithmic}[1]
    \State Initialize $I_{m}$ and $I_{m}^{Aux}$
    \Statex 
    \State \textbf{Client $m$ executes:}
    \State $\mathcal{G}_{m} \leftarrow \left \{ \varnothing \right \}$, $\mathcal{H}_{m} \leftarrow \left \{ \varnothing \right \}$, $u \leftarrow 0$, $v \leftarrow 0$
    \State $\mathcal{G}_{m}\left [ u \right ] \leftarrow w_{g_{m}}$, $\mathcal{H}_{m}\left [ v \right ] \leftarrow w_{h_{m}}$
    \State Build the communication queue $Q_{m}$
    \For{each \textbf{Client} $C_i\in Q_m$ \textbf{in parallel}}
        \State \textbf{Client} $C_i$ sends $I_{i}$, $I_{i}^{Aux}$, $\left | \mathcal{D}_{i} \right |$, and $w_{i}$
        \State Calculate $\text{Div}(m,i)$ by Eq. \eqref{eq:divergence_metric}
        \If{$\text{Div}(m,i)<th_{I}$}
            \State $v \leftarrow v + 1$
            \State $\mathcal{H}_{m}[v] \leftarrow w_{h_{i}}$
        \EndIf
        \State $u \leftarrow u + 1$
        \State $\mathcal{G}_{m}[u] \leftarrow w_{g_{i}}$
    \EndFor
    \State $w_{g_{m}} \leftarrow \sum_{w_{g_u}\in \mathcal{G}_{m}} \frac{|\mathcal{D}_u|}{\sum_{u}|\mathcal{D}_{u}|}w_{g_u}$
    \State $w_{h_{m}} \leftarrow \sum_{w_{h_v}\in \mathcal{H}_{m}} \frac{|\mathcal{D}_v|}{\sum_{v}|\mathcal{D}_{v}|}w_{h_v}$
    \State $w_{m} \leftarrow w_{g_{m}} \bigoplus w_{h_{m}}$
    \State $I^{Aux}_{avg} \leftarrow \frac{\sum_{i}I_{i}^{Aux}+I_{m}^{Aux}}{\left | Q_{m} \right |+1}$
    \State Perform local training by optimizing Eq. \eqref{eq:local_object}
    \State Update $I_{m}$ by Eq. \eqref{eq:std_rep}
    \State Update $I_{m}^{Aux}$ by Eq. \eqref{eq:middle_rep}
\end{algorithmic}
\end{algorithm}

Where $\mathcal{G}_{m}$ denotes a container containing the model parameters of the general feature extractor $w_{g_{m}}$, $\mathcal{H}_{m}$ represents a container containing the model parameters of the personalized classifier $w_{h_{m}}$, $w_{m} := w_{g_{m}} \bigoplus w_{h_{m}}$ is the model parameters on client $m$, and $th_{I}$ is the pre-defined threshold value.

\subsection{Overall Framework of UA-PDFL}
By integrating the method of unit representation with client-wise dropout and layer-wise personalization, we present the overall procedure using pseudo code in Algorithm \ref{algotirhm:UA_PDFL}, where $\mathcal{C}_{M}$ represents all the participating clients in UA-PDFL.

\begin{algorithm}
\caption{UA-PDFL}\label{algotirhm:UA_PDFL}
\begin{algorithmic}[1]
    \State Initialize model parameters  $w_{1:M}=\{ w_{1}^{0}, w_{2}^{0}, \dots,  w_{M}^{0} \}$
    \For {each communication round $r=1, 2, \dots, R$}
        \For {each client $C_i\in \mathcal{C}_{M}$ \textbf{in parallel}}
            \State $\mathcal{G}_{i} \leftarrow \left \{ \varnothing \right \}$, $\mathcal{H}_{i} \leftarrow \left \{ \varnothing \right \}$, $u \leftarrow 0$, $v \leftarrow 0$
            \State $\mathcal{G}_{i}\left [ u \right ] \leftarrow w_{g_{i}}$, $\mathcal{H}_{i}\left [ v \right ] \leftarrow w_{h_{i}}$
            \State Randomly push $N_{\text{com}}$ clients to queue $Q_i$
            \For {each client $C_{j} \in Q_{i}$ \textbf{in parallel}}
                \State Send $I_{j}$ and $I^{Aux}_{j}$ to $C_{i}$
            \EndFor
            \State Compute $\text{Div}(i, j), C_{j} \in Q_{i}$ by Eq.\eqref{eq:divergence_metric}.
            \If{$\forall \text{Div}(i,j) \leq th_{I}$}
                \State Randomly pick $C_{j^{'}}\in Q_i$ \Comment{Client-wise dropout}
                \State $w^{r-1}_{i} \leftarrow w^{r-1}_{j^{'}}$
            \Else
                \For{each client $C_{j} \in Q_{i}$ \textbf{in parallel}}
                    \If{$\text{Div}(i,j)<th_{I}$}
                        \State Send $w_{h_{j}}^{r-1}$ to $C_{i}$
                        \State $v \leftarrow v + 1$, $\mathcal{H}_{i}[v] \leftarrow w_{h_{j}}^{r-1}$
                    \EndIf
                    \State Send $w_{g_{j}}^{r-1}$ to $C_{i}$
                    \State $u \leftarrow u + 1$, $\mathcal{G}_{i}[u] \leftarrow w_{g_{j}}^{r-1}$
                \EndFor
                \State $w_{g_{i}}^{r-1} \leftarrow \sum_{w_{g_u}^{r-1}\in \mathcal{G}_{i}} \frac{|\mathcal{D}_u|}{\sum_{u}|\mathcal{D}_{u}|}w_{g_{u}}^{r-1}$
                \State $w_{h_{i}}^{r-1} \leftarrow \sum_{w_{h_v}^{r-1}\in \mathcal{H}_{i}} \frac{|\mathcal{D}_v|}{\sum_{v}|\mathcal{D}_{v}|}w_{h_v}^{r-1}$
                \State $w_{i}^{r-1} \leftarrow w_{g_{i}}^{r-1} \bigoplus w_{h_{i}}^{r-1}$
            \EndIf
            \State $C_{i}$ \textbf{executes} local updates to get $w_{i}^{r}$ \Comment{Layer-wise personalization}
            \State Update $I_{i}$ by Eq. \eqref{eq:std_rep} \Comment{Unit representation}
            \State Update $I_{i}^{Aux}$ by Eq. \eqref{eq:middle_rep}
        \EndFor
    \EndFor
    \State \textbf{Return} $w_{1:M}=\{ w_{1}^{R}, w_{2}^{R}, \dots,  w_{M}^{R} \}$
    \Statex
    \State \textbf{Client $C_{i}$ executes:}
    \State $I^{Aux}_{avg} \leftarrow \frac{\sum_{j}I_{j}^{Aux}+I_{i}^{Aux}}{\left | Q_{i} \right |+1}$
    \State $w_{i}^{r} \leftarrow w_{i}^{r-1}$
    \For{each epoch $e=1,2,\ldots, E$}
    \For{batch $\mathcal{D}_{B} \in \mathcal{D}_{i}$}
    \State $w_{i}^{r} \leftarrow w_{i}^{r}-\eta\bigtriangledown  \left(\mathcal{L}(w_i^{r};\mathcal{D}_{B}) + \mu\left\|I^{Aux}_i-I^{Aux}_{avg}\right\|_{2}^{2} \right)$
    \EndFor
    \EndFor
    \State \textbf{Return} $w_{i}^{r}$
\end{algorithmic}
\end{algorithm}

At beginning of the training procedure, each client $C_i$ initializes its model parameters $w_{i}^{0}$ and constructs its own communication queue $Q_i$ with $N_{\text{com}}$ randomly connected clients. Subsequently, each $C_{i}$ receives both unit representation $I_{j}$ and auxiliary representation $I_{j}^{Aux}$ from every client $C_{j}$ within queue $Q_i$. It is important to note that, at this stage, model exchanges are not required.

After receiving all the unit representations from $Q_{i}$, each client $C_{i}$ computes the divergence metric of all connected clients with its own unit representation $I_{i}$ using Eq. \eqref{eq:std_rep}. If all the computed metric do not exceed the pre-defined threshold value ($\forall \text{Div}(i,j) \leq th_{I}$), client-wise dropout mechanism would be executed. This mechanism entails replacing the local model parameters of satisfied clients with those of a randomly selected client from the corresponding queue. Otherwise, clients will proceed with layer-wise personalization by querying all the model parameters $w_{g}^{r-1}$ of the general feature extractors. For any client $C_{j} \in Q_{i}$, if $\text{Div}(i,j)<th_{I}$ is satisfied, each $C_{i}$ will query its model parameters $w_{h_{j}}$ of the personalized classifiers. After that, model aggregations and combinations should be performed to generate new $w^{r-1}$ for those clients employing layer-wise personalization (line 20-22 of Algorithm \ref{algotirhm:UA_PDFL}).

Finally, all the clients in DFL execute local training on their local private data $\mathcal{D}_i$ (line 28-32 of Algorithm \ref{algotirhm:UA_PDFL}), and update their unit representation $I$ and auxiliary representation $I^{Aux}$ by Eq. \eqref{eq:std_rep} and Eq. \eqref{eq:middle_rep}, respectively. It should be noted that our proposed UA-PDFL mechanism inherently supports highly efficient parallel computing, and local training and client communication can be operated asynchronously.

\section{Convergence Analysis of UA-PDFL Framework}
In this section, we present a convergence analysis of our proposed UA-PDFL framework, building upon the methodologies and proofs established in the previous work \citep{ma2022like, bertsekas2011incremental}.

\subsection{Assumptions}
\paragraph{Assumption 1} Assume that each local loss function $f_m(w)$ is L-smooth and $\mu$-strongly convex:
\begin{equation}
\label{eq:lusmooth}
\begin{aligned}
    &f_m(w') \leq f_m(w) + \nabla f_m(w)^{\intercal} (w' - w) + \frac{L}{2} \|w' - w\|^2 \\
    &f_m(w') \geq f_m(w) + \nabla f_m(w)^{\intercal} (w' - w) + \frac{\mu}{2} \|w' - w\|^2
\end{aligned}
\end{equation}

\paragraph{Assumption 2} Assume the variance of stochastic gradients across data samples is bounded:
\begin{equation}
    \mathbb{E}\left[\|\nabla L(w; (x, y)) - \nabla f_m(w)\|^2\right] \leq \sigma^2
\end{equation}

\paragraph{Assumption 3} Assume the divergence between two clients $i$ and $j$ is bounded by $\Delta$:
\begin{equation}
    \text{Div}(i, j) \leq \Delta, \quad \forall i, j
\end{equation}

\subsection{Gradient Update Rule}
For $M$ participating clients in DFL, the global objective is to minimize the weighted global loss:
\begin{equation}
    F(w_{1:M}) = \sum_{m=1}^M \alpha_m f_m(w_m)
\end{equation}
where $\alpha_m$ is the aggregation coefficient for client $m$, satisfying $\sum_{m=1}^M \alpha_m = 1$ and the global update rule for $F(w_{1:M})$ is:
\begin{equation}
    w^{r+1} = w^r - \eta \nabla F(w^r) + \eta \xi^r
\end{equation}
where $\xi^{r}$ is the gradient noise introduced by data heterogeneity:
\begin{equation}
    \xi^r = \sum_{m=1}^M \alpha_m \left(\nabla f_m(w_m^r) - \nabla F(w^r)\right)
\end{equation}

By applying client dropout to reduce the influence of similar clients, let the variance of unbiased $\xi^{r}$ bounded as:
\begin{equation}
    \textit{Var}(\xi^{r})=\mathbb{E}[\| \xi^{r} \|^{2}] \leq \frac{\sigma^{2}}{M}
\end{equation}

Furthermore, the auxiliary term reduces divergence across local models, i.e., $\left\|I^{Aux}_m-I^{Aux}_{avg}\right\|_{2}^{2} \rightarrow 0$ as the training round $r \rightarrow \infty$. Therefore, we can omit this part in the following theoretical analysis.

\subsection{Decomposing the Error Using Smoothness}
According to the update $w^{r+1} - w^r = -\eta \nabla F(w^r) + \eta \xi^r$, we can easily get:
\begin{equation}
    \|w^{r+1} - w^r\|^2 = \eta^2 \|\nabla F(w^r)\|^2 - 2\eta^2 \nabla F(w^r)^{\intercal} \xi^r + \eta^2 \|\xi^r\|^2
\end{equation}

Substituting ${w}'=w^{r+1}$ and $w=w^{r}$ into L-smooth assumption of Eq. \eqref{eq:lusmooth} and taking the expectation on both sides of inequality:
\begin{equation}
    \mathbb{E}[F(w^{r+1})] \leq  \mathbb{E}[ F(w^{r})] + \mathbb{E}[ \nabla F(w^{r})^{\intercal} (w^{r+1} - w^{r})] + \frac{L}{2} \mathbb{E}[ \|w^{r+1} - w^{r}\|^2]
\end{equation}

Then, substituting the update rule into $\mathbb{E}[ \nabla F(w^{r})^{\intercal} (w^{r+1} - w^{r})] $, we have:
\begin{equation}
    \mathbb{E}[ \nabla F(w^{r})^{\intercal} (w^{r+1} - w^{r})] = -\eta \mathbb{E}[ \| \nabla F(w^{r}) \|^{2}] -\eta \mathbb{E}[\nabla F(w^{r})^{\intercal} \xi^r]
\end{equation}

If $\xi^{r}$ is unbiased and independent of $\nabla F(w^{r})$, we have $\mathbb{E}[\nabla F(w^{r})^{\intercal} \xi^{r}]=0$, thus:
\begin{equation}
\label{eq:edfw}
     \mathbb{E}[ \nabla F(w^{r})^{\intercal} (w^{r+1} - w^{r})] = -\eta \mathbb{E}[ \| \nabla F(w^{r}) \|^{2}] 
\end{equation}

Similarly, the expectation of $\|w^{r+1} - w^r\|^2$ is:
\begin{equation}
\label{eq:edw}
    \mathbb{E}[\|w^{r+1} - w^r\|^2]=\eta^{2}\mathbb{E}[ \| \nabla F(w^{r}) \|^{2}] + \eta^{2}\mathbb{E}[ \| \xi^{r} \|^{2} ]
\end{equation}

Combing Eq. \eqref{eq:edfw} and Eq. \eqref{eq:edw} into the smoothness inequality, the following formula would be derived:
\begin{equation}
\label{eq:slsmooth}
\begin{aligned}
    \mathbb{E}[F(w^{r+1})] &\leq \mathbb{E}[F(w^r)] - \eta \|\nabla F(w^r)\|^2 + \frac{\eta^2 L}{2} \mathbb{E}[\|\nabla F(w^r)\|^2] + \frac{\eta^2 L}{2} \mathbb{E}[\|\xi^r\|^2] \\
    &\leq \mathbb{E}[F(w^r)] - \eta (1 - \frac{\eta L}{2}) \mathbb{E} [\|\nabla F(w^r)\|^2] + \frac{\eta^2 L}{2} \frac{\sigma^2}{M}
\end{aligned}
\end{equation}

\subsection{Applying Strong Convexity}
For the case of strong convexity satisfying $\|\nabla F(w^r)\|^2 \geq 2\mu (F(w^r) - F(w^*))$ ($w^{*}$ is the minimizer), and letting $\delta^{r}=F(w^r)-F(w^*)$, we have:
\begin{equation}
\begin{aligned}
    \mathbb{E}[\delta^{r+1}] &\leq \mathbb{E}[\delta^{r}] - \eta (1 - \frac{\eta L}{2}) \mathbb{E} [\|\nabla F(w^r)\|^2] + \frac{\eta^2 L}{2} \frac{\sigma^2}{M} \\
    & \leq \mathbb{E}[\delta^{r}] - \eta (1 - \frac{\eta L}{2})2\mu \mathbb{E}[\|\nabla F(w^r)\|^2] + \frac{\eta^2 L}{2} \frac{\sigma^2}{M}
\end{aligned}
\end{equation}

And for small $\eta$, the term $\frac{\eta L}{2}$ is negligible compared to 1. Thus, $ 1 -2\mu \eta (1 - \frac{\eta L}{2})  \approx 1 - \mu \eta $. This approximation simplifies the inequality to:
\begin{equation}
    \mathbb{E}[F(w^{r+1}) - F(w^*)] \leq (1 - \eta \mu) \mathbb{E}[F(w^r) - F(w^*)] + \frac{\eta^2 L}{2} \frac{\sigma^2}{M}
\end{equation}

By iterating the above inequality for $R$ rounds and making $\eta = \frac{1}{L}$, the convergence rate becomes:
\begin{equation}
    \mathbb{E}[F(w^{R}) - F(w^*)] \leq \left(1 - \frac{\mu}{L}\right)^{R} (F(w^0) - F(w^*)) + \frac{\sigma^2}{\mu M}
\end{equation}

\section{Experiments}
To exemplify the viability and robustness of the proposed UA-PDFL framework, extensive experiments were conducted together with several state-of-the-art FL methods. Due to the scarcity of the PDFL method, we modified the communication protocol of some of them. In this section, we first compare the performance of UA-PDFL on image classification tasks across three different datasets. Subsequently, we present a case study to validate the effectiveness of standard representation. Finally, an ablation study is conducted.

\subsection{Experimental Settings}

\subsubsection{Models}
To simulate and evaluate FL system across various scales, three levels of neural network models are utilized. The first is a simple convolutional neural network (CNN) comprising two convolutional layers and two fully connected classification layers, which mirrors the structure of our previous work \citep{fedtsdp}. This CNN model is introduced to simulate a light-weight FL system. The second model is ResNet18 \citep{he2016deep}, known for its residual connections that address the vanishing gradient problem in deep neural networks. ResNet18 is adopted to simulate a FL system with relatively heavier computational costs. Lastly, we incorporate VGG11 \citep{simonyan2014very}, a model renowned for its heavy memory requirements due to its larger parameter count compared to the other two models. VGG11 is utilized to simulate a FL system with significant memory costs.

To further divide the models into a feature extractor $g(\cdot)$ and a personalized classifier $h(\cdot)$, we assume that all convolutional layers constitute $g(\cdot)$, while other layers (mostly fully connected layers) constitute $h(\cdot)$. Specifically, the initial 2, 60, and 28 layers of CNN, ResNet18, and VGG11, respectively, are defined as $g(\cdot)$, while the remaining layers are designated as $h(\cdot)$.

\subsubsection{Datasets}
To simulate tasks with varying difficulties, four benchmark image classification datasets, CIFAR10, CIFAR100 \citep{cifar10}, and SVHN \citep{svhn} along with a real-world dataset, PathMNIST \citep{medmnistv2} are adopted. CIFAR10 offers a moderately challenging training task with its ten distinct object classes. Conversely, CIFAR100 presents an exceptionally challenging task by encompassing 100 diverse object classes, thereby posing significant classification challenges. In contrast, SVHN represents a comparatively easier task, comprising digits ranging from 0 to 9, which are thus relatively straightforward to classify. Finally, PathMNIST comprises over 100,000 real-world images, representing 9 distinct categories of histological images of human colorectal cancer and healthy tissue. These images are sourced from various clinical centers, providing a diverse and comprehensive dataset for medical image analysis.

\begin{figure}[hbtp]
    \centering
    \subcaptionbox{$\beta=0.5$}{\includegraphics[width=0.45\textwidth]{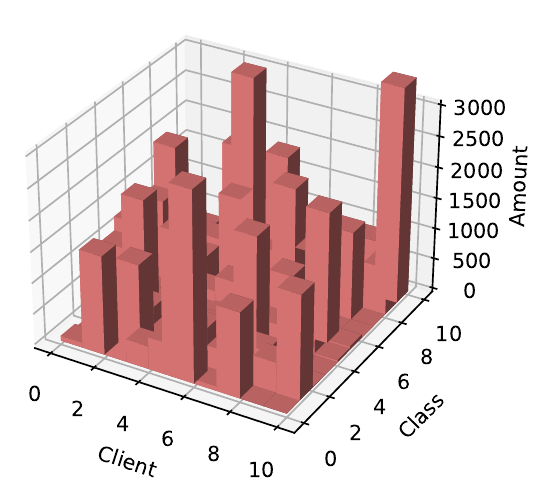}}
    \hfill
    \subcaptionbox{$\beta=5$}{\includegraphics[width=0.45\textwidth]{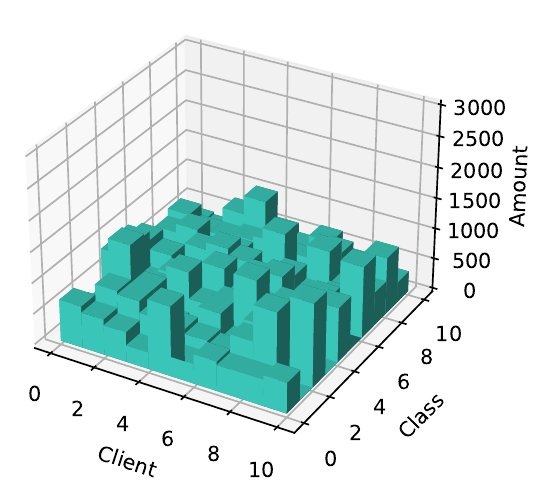}}
\caption{Illustration of client data distributions with different $\beta$ values.}
\label{fig:distribution-dir}
\end{figure}

\begin{figure}[hbtp]
    \centering
    \includegraphics[width=0.46\textwidth]{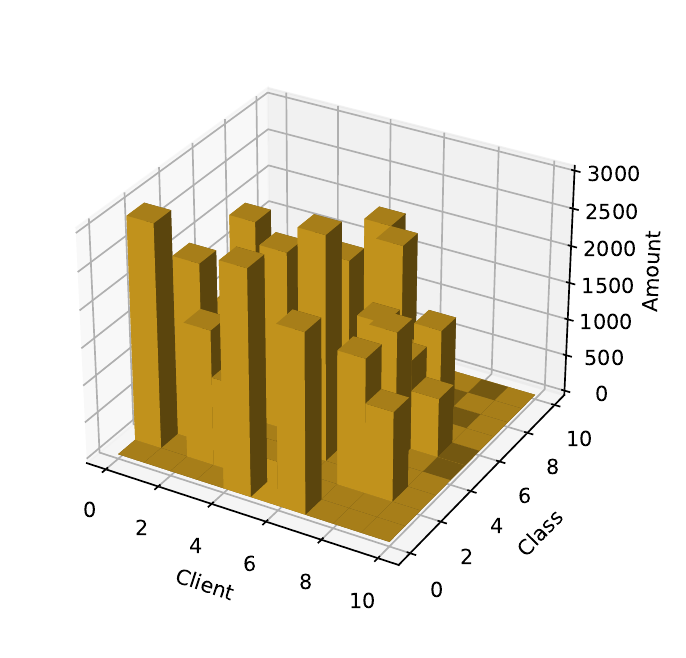}
\caption{Illustration of client data with extreme cross-domain distribution.}
\label{fig:distribution-noniid}
\end{figure}

To simulate non-IID data reflecting the real-world scenarios, each client is assigned a proportion of training data for each label class based on the Dirichlet distribution \citep{moon}. The hyperparameter $\beta \sim \text{Dir}(\beta)$ governs governs the degree of heterogeneity in the allocation. In our experiments, we use $\beta=0.5$ and $\beta=5$ to simulate highly non-IID and slightly non-IID data, respectively. A smaller $\beta$ results in a more imbalanced data partition. To further simulate extreme cross-domain non-IID data for the real-world data experiments, we partitioned the dataset following the method outlined in \citep{fedavg}. Each client is assigned samples from only a few classes, with significant non-overlap between the samples across clients. The specific partitioning results are illustrated in Fig. \ref{fig:distribution-noniid}.

\subsubsection{Algorithms Under Comparisons}
In the experiments, we benchmark our proposed method against the following baseline approaches:

\begin{itemize}
    \item Local: Training is locally performed without communication among clients.
    \item FedAvg \citep{fedavg}: The vanilla FL method conducts weighted averaging directly without any intermediate aggregation mechanisms.
    \item CriticalFL \citep{yan2023criticalfl}: Adjust the number of client communications according to gradient information. This serves as the baseline for communication efficiency. 
    \item FedPer \citep{fedper}: The first to introduce personalized layers into FL, serving as a classical PFL baseline.
    \item DisPFL \citep{dai2022dispfl}: A recent state-of-the-art algorithm in PDFL utilizes decentralized sparse training to achieve a personalized model.
    \item DFedAvgM \citep{9850408}: A recent advancement in DFL incorporates heavy-ball momentum to mitigate the need for frequent communication, thereby improving efficiency.
\end{itemize}

Please note that FedAvg, CriticalFL and FedPer are originally performed on centralized FL environments. To ensure more fair comparisons, we have modified their communication protocols. Generally, each client servers as a central server itself, asynchronously collecting model from other clients. This mirrors the pattern outlined in Algorithm \ref{algotirhm:UA_PDFL}. 

\subsubsection{Other Settings}
All other training setups for UA-PDFL are listed as follows:

\begin{itemize}
    \item Total number of clients $|C|$: 30
    \item Total number of communication rounds $R$: 150
    \item Number of communicated clients $N_{\text{com}}$: 5
    \item Initial learning rate $\eta$: 0.05
    \item Leaning momentum: 0.5
    \item Learning rate decay: 0.95
    \item Training batch size: 50
\end{itemize}

\subsection{UA-PDFL VS Baseline Approaches}
In this section, we verify the effectiveness of our proposed UA-PDFL across varying degrees of data heterogeneity. The learning performance of UA-PDFL is compared with the aforementioned five baseline approaches.

\begin{table}[hbtp]
\caption{Final Test Accuracy with $\beta=0.5$}
\label{tab:exp-dir0.5}
\resizebox{\textwidth}{!}{
\begin{tabular}{l|lllllllll}
    \hline
        \diagbox{Dataset}{Algorithm} & Local & FedAvg & CriticalFL & FedPer & DisPFL & DFedAvgM & Proposed \\
    \hline
        \multicolumn{8}{c}{CNN} \\
    \hline
        CIFAR10 & 65.06\%$\pm$2.06 & 67.16\%$\pm$0.44 & 66.98\%$\pm$0.04 & 74.67\%$\pm$0.41 & 73.57\%$\pm$0.44 & 69.11\%$\pm$0.10 & \textbf{74.85\%$\pm$0.95} \\
        CIFAR100 & 29.29\%$\pm$0.01 & 34.47\%$\pm$0.02 & 34.10\%$\pm$0.37 & 30.66\%$\pm$0.11 & 32.15\%$\pm$0.54 & 35.36\%$\pm$0.10 & \textbf{36.09\%$\pm$0.06} \\
        SVHN & 84.85\%$\pm$0.40 & 88.68\%$\pm$0.09 & 87.84\%$\pm$0.60 & \textbf{91.28\%$\pm$0.08} & 82.61\%$\pm$0.64 & 89.93\%$\pm$0.02 & 90.79\%$\pm$0.03 \\
    \hline
        \multicolumn{8}{c}{ResNet18} \\
    \hline
        CIFAR10 & 67.43\%$\pm$1.34 & 71.18\%$\pm$0.20 & 72.90\%$\pm$0.84 & 79.21\%$\pm$0.30 & 77.34\%$\pm$0.73 & 71.17\%$\pm$0.64 & \textbf{79.66\%$\pm$0.10} \\
        CIFAR100 & 30.18\%$\pm$0.06 & 35.30\%$\pm$0.15 & 36.93\%$\pm$0.08 & 39.90\%$\pm$0.15 & 33.76\%$\pm$0.26 & 34.49\%$\pm$0.54 & \textbf{40.23\%$\pm$0.09} \\
        SVHN & 87.56\%$\pm$0.06 & 92.99\%$\pm$0.06 & 93.18\%$\pm$0.07 & 93.10\%$\pm$0.83 & \textbf{94.37\%$\pm$0.23} & 93.29\%$\pm$0.03 & 93.80\%$\pm$0.33 \\
    \hline
        \multicolumn{8}{c}{VGG11} \\
    \hline
        CIFAR10 & 70.53\%$\pm$0.59 & 74.96\%$\pm$0.92 & 75.30\%$\pm$0.38 & \textbf{80.50\%$\pm$0.23} & 79.61\%$\pm$0.23 & 75.44\%$\pm$0.29 & 80.00\%$\pm$0.28 \\
        CIFAR100 & 29.31\%$\pm$0.07 & 30.78\%$\pm$0.29 & 27.50\%$\pm$1.14 & 43.61\%$\pm$0.78 & 34.84\%$\pm$0.08 & 33.63\%$\pm$0.01 & \textbf{46.08\%$\pm$0.39} \\
        SVHN & 86.93\%$\pm$0.40 & 92.23\%$\pm$0.01 & 92.04\%$\pm$0.09 & 94.11\%$\pm$0.02 & 93.01\%$\pm$0.17 & 92.27\%$\pm$0.03 & \textbf{94.80\%$\pm$0.01} \\
    \hline
\end{tabular}
}
\end{table}

\begin{figure}[hbtp]
    \centering
    \includegraphics[width=0.92\textwidth]{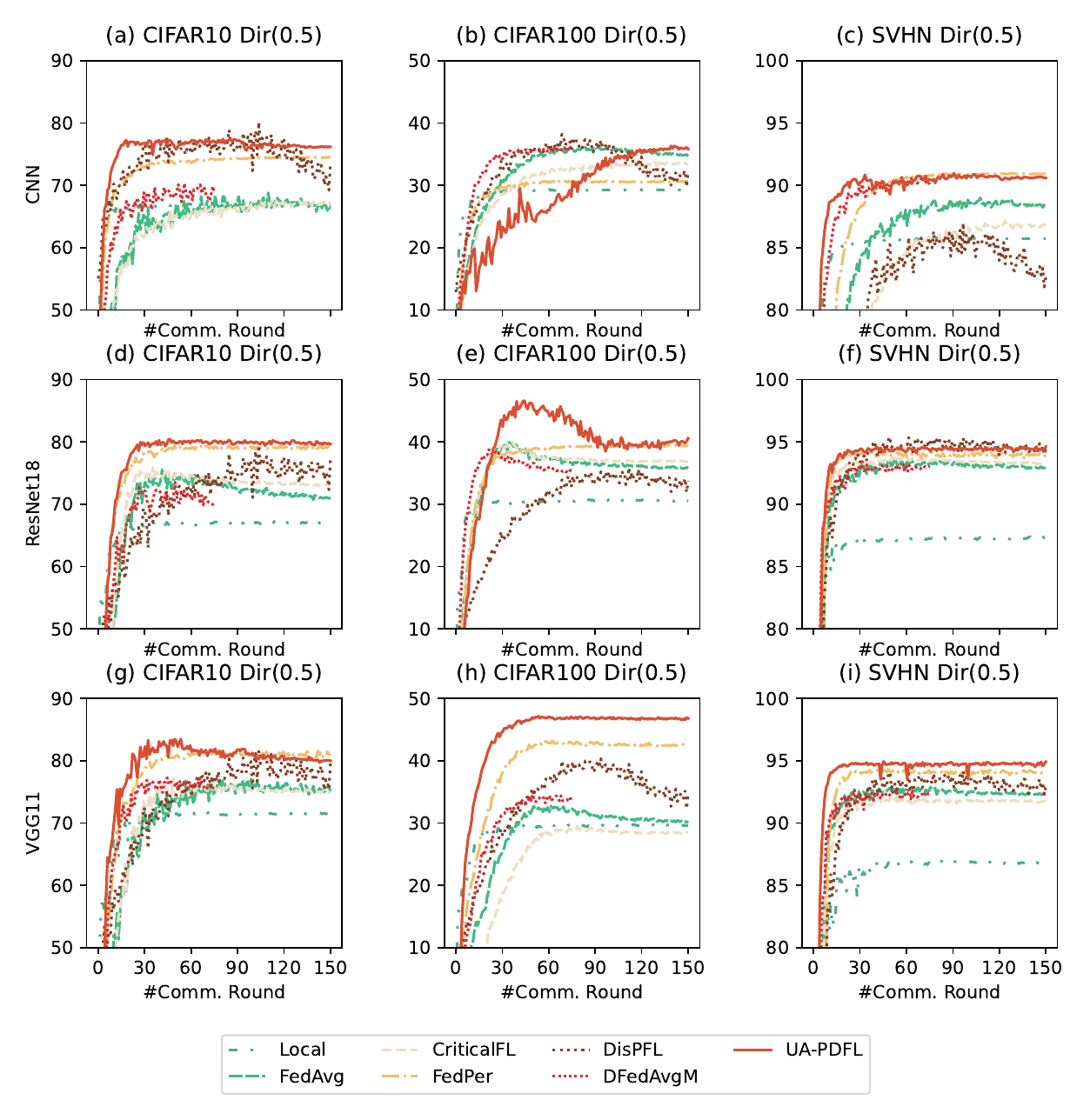}
\caption{The test accuracy over communication rounds with $\beta=0.5$.}
\label{fig:convergence-0.5}
\end{figure}

The final test accuracy averaged across participating clients on highly heterogeneous data ($\beta=0.5$) are presented in Table \ref{tab:exp-dir0.5}. It is evident that the proposed UA-PDFL outperforms other methods in most experiments, especially on CIFAR100 with an accuracy of approximate 40.03\% for ResNet18 model. However, the performance of UA-PDFL is slightly lower than FedPer and DisPFL in a few cases, but by no more than 1\%. In addition, it can be inferred that the incorporation of personalization layers can yield substantial performance enhancements, particularly in highly heterogeneous data environments. Except that, the real-time learning performance across participating clients over communication rounds are shown in Fig. \ref{fig:convergence-0.5}. It is obvious that our proposed UA-PDFL (solid line) demonstrates a notably faster convergence speed compared to other baseline approaches across all the three datasets with $\text{Dir}(0.5)$. Notably, DFedAvgM emphasizes increased local training iterations, and in comparative experiments, its local training iterations were set to twice those of other algorithms. To ensure fairness, the number of communication rounds for DFedAvgM was correspondingly reduced to half, amounting to 75 communication rounds.

\begin{table}[hbtp]
\caption{Final Test Accuracy with $\beta=5$}
\label{tab:exp-dir5}
\resizebox{\textwidth}{!}{
\begin{tabular}{l|lllllll}
    \hline
        \diagbox{Dataset}{Algorithm} & Local & FedAvg & CriticalFL & FedPer & DisPFL & DFedAvgM & Proposed \\
    \hline
        \multicolumn{8}{c}{CNN} \\
    \hline
        CIFAR10 & 49.13\%$\pm$0.24 & 69.38\%$\pm$0.14 & 69.44\%$\pm$0.70 & 61.85\%$\pm$0.26 & 65.93\%$\pm$0.21 & 69.57\%$\pm$0.28 &  \textbf{71.34\%$\pm$0.03} \\ 
        CIFAR100 & 14.84\%$\pm$0.06 & 30.83\%$\pm$0.24 & 30.13\%$\pm$1.74 & 14.06\%$\pm$0.08 & 28.90\%$\pm$0.62 & 33.88\%$\pm$0.23 & \textbf{36.83\%$\pm$0.25} \\ 
        SVHN & 79.81\%$\pm$0.44 & 89.67\%$\pm$0.07 & 89.53\%$\pm$0.08 & 88.33\%$\pm$0.03 & 78.90\%$\pm$0.18 & 90.46\%$\pm$0.02 & \textbf{91.12\%$\pm$0.01} \\ 
    \hline
        \multicolumn{8}{c}{ResNet18} \\
    \hline
        CIFAR10 & 48.58\%$\pm$0.38 & 75.79\%$\pm$0.32 & \textbf{77.84\%$\pm$0.01} & 77.33\%$\pm$0.03 & 66.29\%$\pm$0.39 & 75.31\%$\pm$0.20 & 77.24\%$\pm$0.11 \\
        CIFAR100 & 15.72\%$\pm$0.10 & 35.78\%$\pm$0.28 & 37.07\%$\pm$0.11 & 24.23\%$\pm$0.07 & 27.77\%$\pm$0.21 & 33.50\%$\pm$0.46 & \textbf{40.03\%$\pm$0.31} \\
        SVHN & 83.83\%$\pm$0.02 & 93.32\%$\pm$0.04 & 94.00\%$\pm$0.04 & 93.26\%$\pm$0.01 & 92.91\%$\pm$0.37 & 93.50\%$\pm$0.00 & \textbf{93.64\%$\pm$0.05} \\
    \hline
        \multicolumn{8}{c}{VGG11} \\
    \hline
        CIFAR10 & 55.76\%$\pm$0.09 & 78.12\%$\pm$0.22 & 78.65\%$\pm$0.23 & 79.37\%$\pm$0.02 & 71.21\%$\pm$0.14 & 78.19\%$\pm$0.12 & \textbf{79.76\%$\pm$0.01} \\
        CIFAR100 & 14.92\%$\pm$0.17 & 27.19\%$\pm$0.14 & 29.20\%$\pm$1.13 & 29.72\%$\pm$0.77 & 31.74\%$\pm$0.10 & 32.77\%$\pm$0.11 & \textbf{32.99\%$\pm$0.17} \\
        SVHN & 83.80\%$\pm$0.12 & 92.76\%$\pm$0.01 & 92.73\%$\pm$0.06 & 93.25\%$\pm$0.00 & 91.36\%$\pm$0.50 & 92.98\%$\pm$0.06 & \textbf{93.52\%$\pm$0.00} \\
    \hline
\end{tabular}
}
\end{table}

\begin{figure}[hbtp]
    \centering
    \includegraphics[width=0.92\textwidth]{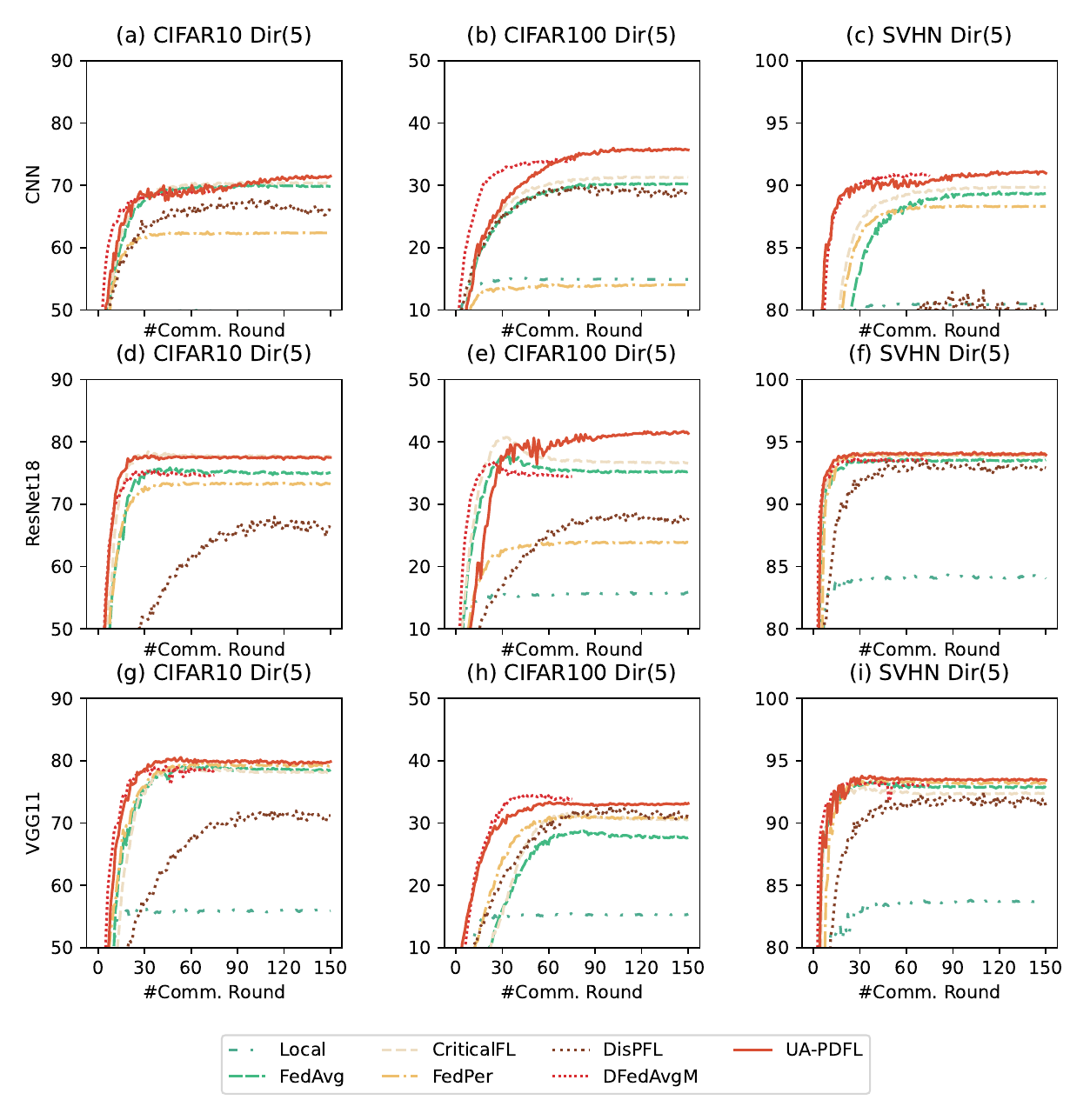}
\caption{The test accuracy over communication rounds with $\beta=5$.}
\label{fig:convergence-5}
\end{figure}

The outcomes of the final test accuracy on relatively low heterogeneous data ($\beta=5$) are shown in Table \ref{tab:exp-dir5}. In such a scenario, clients in DFL would share similar local learning objectives. Notably, UA-PDFL demonstrates a promising performance for all scenarios, just trailing by 0.6\% compared to CriticalFL when employing the ResNet18 model on CIFAR10 dataset. It is interesting to note that the incorporation of personalization layers does not appear to yield improvements in enhancing learning performance for relatively IID data, in contrast to the outcomes observed for relatively non-IID data. This phenomenon can be attributed to the fact that most PFL methods maintain a constant number of personalization layers across varying degrees of data skew, resulting in poor learning performance on data with low heterogeneity. In addition, the test accuracy over communication rounds with $\beta=5$ are depicted in Fig. \eqref{fig:convergence-5}, and our proposed UA-PDFL (solid lines) converges much faster than other five baseline approaches across most cases.

From a dataset perspective, the training performance on the relatively challenging task (CIFAR100) experiences a substantial boost. This enhancement can be attributed to the alignment of the feature extractor (Eq. \eqref{eq:local_object}). Conversely, for easier tasks such as SVHN, training a general feature extractor is comparatively less challenging, resulting in a relatively lower enhancement in learning performance.

\subsection{The Number of Communicated Clients}
Considering the critical importance of communication efficiency in DFL, this section focuses on evaluating the performance of the proposed algorithm in this context. By varying the number of communicating clients, $N_{\text{com}}$, different levels of communication overhead in DFL are simulated, allowing for an analysis of their impact on the model's final performance.

\begin{table}[hbtp]
\caption{Final Test Accuracy with $\beta=0.5$}
\label{tab:selection-0.5}
\resizebox{\textwidth}{!}{
\begin{tabular}{l|llll}
    \hline
        \diagbox{Dataset}{$N_{com}$} & 1 & 5 & 15 & 29 \\
    \hline
        \multicolumn{4}{c}{CNN} \\
    \hline
        CIFAR10 & 72.21\%$\pm$0.81 & 74.85\%$\pm$0.95 & 73.51\%$\pm$0.37 & 73.04\%$\pm$0.50 \\ 
        CIFAR100 & 36.29\%$\pm$0.45 & 36.09\%$\pm$0.06  & 35.96\%$\pm$0.22 & 35.91\%$\pm$0.39 \\ 
        SVHN & 89.96\%$\pm$0.39 & 90.79\%$\pm$0.03  & 90.85\%$\pm$0.07 & 90.73\%$\pm$0.09 \\ 
    \hline
        \multicolumn{4}{c}{ResNet18} \\
    \hline  
        CIFAR10 & 75.59\%$\pm$1.05 & 79.66\%$\pm$0.10 & 80.04\%$\pm$0.62 & 80.93\%$\pm$1.10 \\
        CIFAR100 & 33.17\%$\pm$0.80 & 40.23\%$\pm$0.09 & 38.38\%$\pm$0.14 & 38.54\%$\pm$0.10 \\
        SVHN & 93.44\%$\pm$0.09 & 93.80\%$\pm$0.33 & 93.79\%$\pm$0.04 & 94.06\%$\pm$0.15 \\
    \hline
        \multicolumn{4}{c}{VGG11} \\
    \hline
        CIFAR10 & 77.32\%$\pm$2.07 & 80.00\%$\pm$0.28 & 79.96\%$\pm$0.00 & 79.37\%$\pm$0.51 \\
        CIFAR100 & 38.94\%$\pm$1.24 & 46.08\%$\pm$0.39 & 46.92\%$\pm$0.36 & 46.95\%$\pm$2.15 \\
        SVHN & 93.97\%$\pm$0.04 & 94.80\%$\pm$0.01 & 94.65\%$\pm$0.00 & 94.54\%$\pm$0.15 \\
    \hline
\end{tabular}
}
\end{table}

The final training outcomes on data characterized by a relatively skewed distribution ($\beta=0.5$) are detailed in Table \ref{tab:selection-0.5}. Notably, it is evident that the training performance exhibits a tendency towards poorer outcomes when the number of connections is set to 1. With an increase in the number of connections, there is a discernible, albeit gradual, enhancement in training performance. However, this amelioration appears to reach a plateau, notably observed when the number of connections surpasses 5. Beyond this threshold, augmenting the number of clients engaged in training does not yield a substantial enhancement in performance.

\begin{table}[hbtp]
\caption{Final Test Accuracy with $\beta=5$}
\label{tab:selection-5}
\resizebox{\textwidth}{!}{
\begin{tabular}{l|llll}
    \hline
        \diagbox{Dataset}{$N_{com}$} & 1 & 5 & 15 & 29 \\
    \hline
        \multicolumn{4}{c}{CNN} \\
    \hline
        CIFAR10 & 71.87\%$\pm$0.16 & 71.34\%$\pm$0.03 & 71.35\%$\pm$0.93 & 71.49\%$\pm$0.03 \\ 
        CIFAR100 & 37.74\%$\pm$1.31 & 36.83\%$\pm$0.25 & 38.18\%$\pm$0.17 & 36.74\%$\pm$0.17 \\ 
        SVHN & 91.08\%$\pm$0.10 & 91.12\%$\pm$0.01 & 90.99\%$\pm$0.01 & 91.04\%$\pm$0.01 \\ 
    \hline
        \multicolumn{4}{c}{ResNet18} \\
    \hline  
        CIFAR10 & 70.96\%$\pm$2.45 & 77.24\%$\pm$0.11 & 78.37\%$\pm$0.02 & 78.85\%$\pm$0.01 \\
        CIFAR100 & 42.38\%$\pm$0.36 & 40.03\%$\pm$0.31 & 38.99\%$\pm$0.43 & 38.06\%$\pm$0.49 \\
        SVHN & 93.38\%$\pm$0.03 & 93.64\%$\pm$0.05 & 93.73\%$\pm$0.14 & 94.03\%$\pm$0.01 \\
    \hline
        \multicolumn{4}{c}{VGG11} \\
    \hline
        CIFAR10 & 79.08\%$\pm$0.08 & 79.76\%$\pm$0.01 & 79.78\%$\pm$0.07 & 79.85\%$\pm$0.03 \\
        CIFAR100 & 25.35\%$\pm$0.34 & 32.99\%$\pm$0.17 & 32.93\%$\pm$0.28 & 33.79\%$\pm$0.11 \\
        SVHN & 93.07\%$\pm$0.07 & 93.52\%$\pm$0.00 & 93.68\%$\pm$0.01 & 93.60\%$\pm$0.06 \\
    \hline
\end{tabular}
}
\end{table}

On the other hand, the final test accuracy achieved under a relatively homogeneous distribution ($\beta=5$) are outlined in Table \ref{tab:selection-5}. Consistent with previous observations, it is evident that greater client involvement in training correlates with improved training performance. However, a noteworthy anomaly arises when training the ResNet18 classification model on the demanding CIFAR100 dataset. Specifically, an inverse relationship is observed between the number of participating clients and training performance. This anomaly may be attributed to the architectural features of ResNet18, particularly its employment of shortcut connections. Furthermore, from a broader model perspective, it is apparent that the performance gains stemming from augmenting the number of training clients are not substantial, particularly given the relatively fewer parameters inherent in CNN models. This trend persists even across simpler classification datasets, such as SVHN.

\begin{figure}[hbtp]
    \centering
    \includegraphics[width=0.92\textwidth]{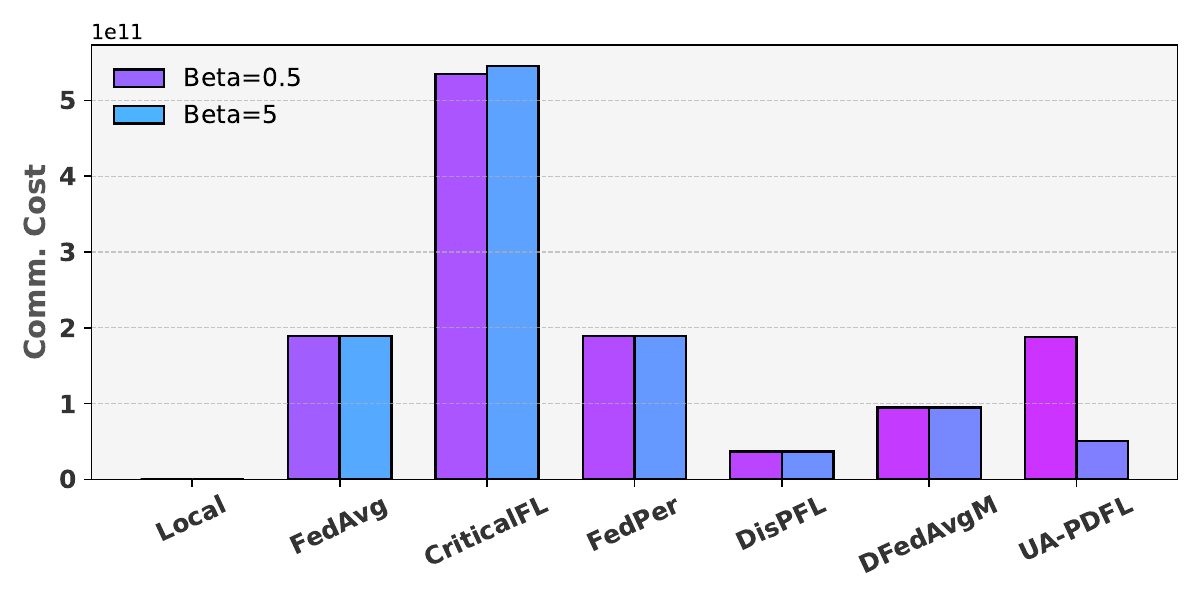}
\caption{Communication overhead across algorithms under varying data distributions ($\beta=0.5$ and $\beta=5$).}
\label{fig:comm}
\end{figure}

In the experiments above, the total parameter transmission volume is illustrated in Fig. \ref{fig:comm}. As observed, UA-PDFL does not significantly reduce communication overhead compared to traditional methods when the data distribution is highly uneven. However, when the data distribution is more uniform, client-wise dropout mechanism leads to a notable reduction in communication volume. While the communication overhead of UA-PDFL is higher than that of DFedAvgM and DisPFL, the additional communication effectively captures the features of local data, resulting in improved model accuracy. This trade-off proves advantageous for model training.

\subsection{Simulation Results on the Real-World Dataset}
\label{sec:real-world}
In this section, we evaluate our proposed algorithm alongside baseline methods using a real-world dataset, PathMNIST \citep{medmnistv2}. The corresponding simulation results for both extreme non-IID and IID data are shown in Table \ref{tab:real-world}. It can be observed that on the real-world dataset, UA-PDFL significantly outperforms FedAvg, CriticalFL, DisPFL, and DFedAvgM for non-IID data, while achieving performance comparable to FedPer with final accuracy of 93.52\%$\pm$2.92 and 92.01\%$\pm$1.78 using CNN and ResNet18, respectively. This is because, in scenarios with uneven data distribution, the overlap of generalized features across datasets is limited. Exchanging fewer parameters to preserve local features is more beneficial for federated training. And both FedPer and UA-PDFL adopt layer-wise parameter exchange techniques, which provide a natural training advantage in such scenarios.

\begin{table}[hbtp]
\caption{Simulation Results on PathMNIST.}
\label{tab:real-world}
\resizebox{\textwidth}{!}{
\begin{tabular}{l|llllll}
    \hline
        Model & FedAvg & CriticalFL & FedPer & DisPFL & DFedAvgM & Proposed \\
    \hline
        \multicolumn{7}{c}{Extreme non-IID} \\
    \hline
         CNN & 76.40\%$\pm$1.21 & 72.44\%$\pm$0.92 & 95.28\%$\pm$0.56 & 79.80\%$\pm$0.23 & 77.79\%$\pm$2.24 & 93.52\%$\pm$2.92 \\ 
         ResNet18 & 63.65\%$\pm$2.86 & 59.98\%$\pm$2.98 & 91.05\%$\pm$1.20 & 86.64\%$\pm$0.72 & 68.50\%$\pm$1.85 & 92.01\%$\pm$1.78 \\
    \hline
        \multicolumn{7}{c}{IID} \\
    \hline
        CNN & 85.76\%$\pm$0.05 & 86.43\%$\pm$0.13 & 83.77\%$\pm$0.24 & 70.25\%$\pm$0.17 & 89.40\%$\pm$0.09 & 86.23\%$\pm$0.32 \\
        ResNet18 & 94.55\%$\pm$0.03 & 95.35\%$\pm$0.12 & 95.47\%$\pm$0.03 & 75.98\%$\pm$0.32 & 93.85\%$\pm$0.00 & 96.07\%$\pm$0.08 \\
    \hline
\end{tabular}
}
\end{table}

However, In the case of IID data, FedPer exhibits a significant performance decline, achieving a test accuracy of only 83.77\%$\pm$0.24 on CNN. While UA-PDFL dynamically adjusts the shared layers of model parameters, allowing it to effectively adapt to relatively uniform data distributions. This highlights the robustness of the proposed UA-PDFL in handling varying degrees of data distribution in DFL.

\subsection{Ablation Study on Unit Representation}
\label{sec:analysis_std_rep}
\begin{figure}[hbtp]
    \centering
    \includegraphics[width=0.92\textwidth]{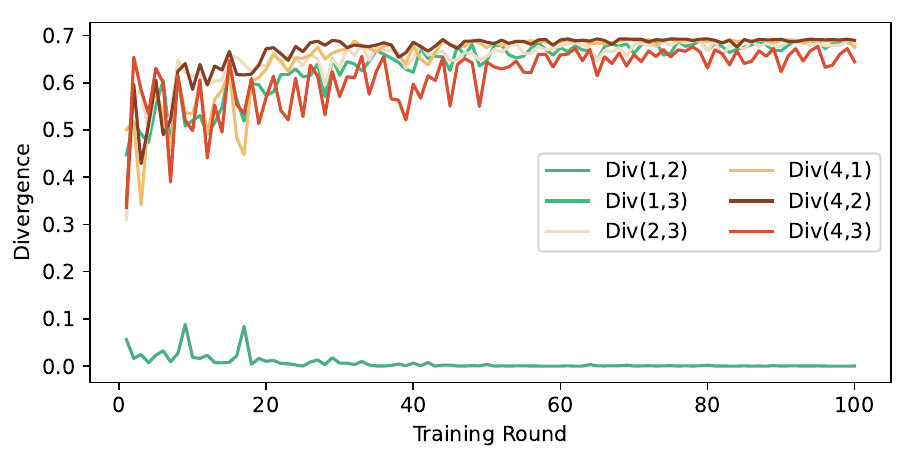}
\caption{The divergence metric over training rounds. Each line represents the divergence metric between two clients.}
\label{fig:analysis_std_rep}
\end{figure}

To further verify the effectiveness of the proposed unit representation, we conduct additional experiments outlined in Section \ref{sec:std_rep}. The results of divergence metric between two clients are shown in Fig. \ref{fig:analysis_std_rep}. It is evident that all the metrics undergo significant fluctuations at the early stage of the training rounds. Specifically, client 1 and 2, sharing identical data distributions, exhibit convergence towards a common direction, resulting in a relatively low divergence metric. Conversely, client 3 and 4, with a distinct data distribution, yields a comparatively higher divergence metric. Moreover, the divergence metric stabilizes around the 30th epoch.

\begin{figure}[hbtp]
    \centering
    \subcaptionbox{$\beta=0.5$}{\includegraphics[width=0.46\textwidth]{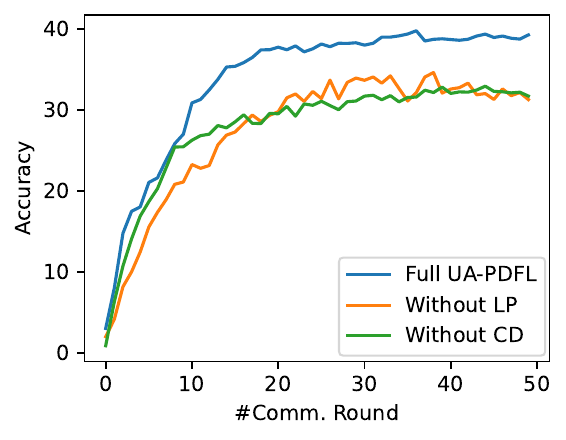}}
    \hfill
    \subcaptionbox{$\beta=5$}{\includegraphics[width=0.46\textwidth]{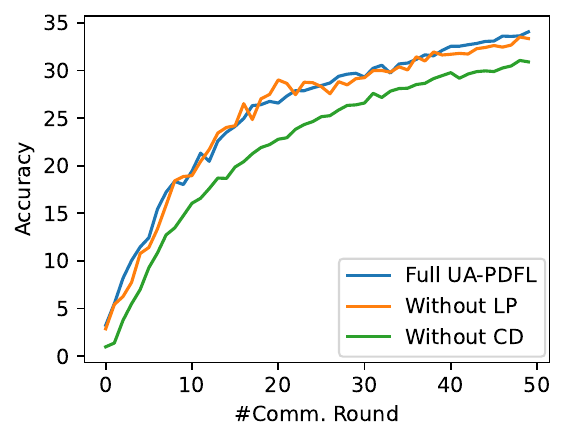}}
\caption{Test accuracy with or without layer-wise personalization or client-wise dropout, where LP denotes Layer-wise Personalization, CD denotes client-wise dropout.}
\label{fig:ablation}
\end{figure}

Furthermore, the impact of client-wise dropout and layer-wise personalization on learning performance, evaluated based on unit representation, is depicted in Fig. \ref{fig:ablation}. To evaluate their individual contributions to UA-PDFL performance, these mechanisms are applied independently. For relatively non-IID data, both layer-wise personalization (Without CD) and client-wise dropout (Without LP) significantly enhance learning performance. Furthermore, the combination of these two mechanisms (Full UA-PDFL) achieves the best overall performance. In scenarios with more balanced data distributions, Full UA-PDFL performs similarly to Without LP, suggesting that client-wise dropout serves as the primary factor, achieving performance comparable to its standalone application.

\section{Conclusion}
In this paper, we explore scenarios within the DFL paradigm, where the coordination of a central server is absent. To address the challenge of non-IID data prevalent in DFL, we introduce a novel Unit representation Aided Personalized DFL framework called UA-PDFL. Unlike approaches relying on centralized public data, UA-PDFL leverages unit representation to discern local data distributions, facilitating the measurement of divergence metrics among clients. Moreover, we employ a client-wise dropout mechanism to alleviate overfitting issues in cases of relatively IID client data. Additionally, a layer-wise personalization technique is proposed to guide the general feature extractor alongside a personalized classifier. This approach enables clients to enhance their feature extraction capabilities while retaining local classification biases.

Extensive experiments have been conducted to compare our proposed UA-PDFL with five baseline DFL methods. The results showcase that our method yields learning performance that is superior or comparable to others, demonstrating its effectiveness across varying levels of data heterogeneity. This advantage can be attributed to the client-wise dropout mechanism, where a randomly selected client model is utilized to replace the local model, particularly beneficial for homogeneous data scenarios. Additionally, the adoption of layer-wise personalization enables the adaptive maintenance of personalized features within the local classifier, particularly advantageous for relatively non-IID data distributions.

The current study constitutes a crucial initial stride towards integrating DFL with personalized FL techniques. Despite the promising empirical outcomes attained, it is noteworthy that the communication costs incurred by DFL are evidently higher compared to those of vanilla FL. In forthcoming research endeavors, our focus will be on refining DFL algorithms to bolster the communication efficiency of the DFL framework.

\section*{CRediT authorship contribution statement}
\textbf{Hangyu Zhu:}Methodology, Writing - Review \& Editing, Supervision.
\textbf{Yuxiang Fan:}Conceptualization, Methodology, Software, Writing - Original Draft.
\textbf{Zhenping Xie:}Supervision, Project administration, Funding acquisition.

\section*{Acknowledgements}
This work was supported in part by the National Natural Science Foundation of China (NSFC) under Grant 62272201, and in part by the Wuxi Science and Technology Development Fund Project under Grant K20231012.

\bibliographystyle{elsarticle-harv} 
\bibliography{ref}

\begin{thebibliography}{63}
\expandafter\ifx\csname natexlab\endcsname\relax\def\natexlab#1{#1}\fi
\providecommand{\url}[1]{\texttt{#1}}
\providecommand{\href}[2]{#2}
\providecommand{\path}[1]{#1}
\providecommand{\DOIprefix}{doi:}
\providecommand{\ArXivprefix}{arXiv:}
\providecommand{\URLprefix}{URL: }
\providecommand{\Pubmedprefix}{pmid:}
\providecommand{\doi}[1]{\href{http://dx.doi.org/#1}{\path{#1}}}
\providecommand{\Pubmed}[1]{\href{pmid:#1}{\path{#1}}}
\providecommand{\bibinfo}[2]{#2}
\ifx\xfnm\relax \def\xfnm[#1]{\unskip,\space#1}\fi
\bibitem[{Arivazhagan et~al.(2019)Arivazhagan, Aggarwal, Singh and Choudhary}]{fedper}
\bibinfo{author}{Arivazhagan, M.G.}, \bibinfo{author}{Aggarwal, V.}, \bibinfo{author}{Singh, A.K.}, \bibinfo{author}{Choudhary, S.}, \bibinfo{year}{2019}.
\newblock \bibinfo{title}{Federated learning with personalization layers}.
\newblock \bibinfo{journal}{arXiv preprint arXiv:1912.00818} .
\bibitem[{Bertsekas(2011)}]{bertsekas2011incremental}
\bibinfo{author}{Bertsekas, D.P.}, \bibinfo{year}{2011}.
\newblock \bibinfo{title}{Incremental gradient, subgradient, and proximal methods for convex optimization: A survey} .
\bibitem[{Briggs et~al.(2020)Briggs, Fan and Andras}]{flhc}
\bibinfo{author}{Briggs, C.}, \bibinfo{author}{Fan, Z.}, \bibinfo{author}{Andras, P.}, \bibinfo{year}{2020}.
\newblock \bibinfo{title}{Federated learning with hierarchical clustering of local updates to improve training on non-iid data}, in: \bibinfo{booktitle}{2020 International Joint Conference on Neural Networks (IJCNN)}, \bibinfo{organization}{IEEE}. pp. \bibinfo{pages}{1--9}.
\bibitem[{Cho et~al.(2023)Cho, Wang, Chirvolu and Joshi}]{comet}
\bibinfo{author}{Cho, Y.J.}, \bibinfo{author}{Wang, J.}, \bibinfo{author}{Chirvolu, T.}, \bibinfo{author}{Joshi, G.}, \bibinfo{year}{2023}.
\newblock \bibinfo{title}{Communication-efficient and model-heterogeneous personalized federated learning via clustered knowledge transfer}.
\newblock \bibinfo{journal}{IEEE Journal of Selected Topics in Signal Processing} \bibinfo{volume}{17}, \bibinfo{pages}{234--247}.
\bibitem[{Dai et~al.(2022)Dai, Shen, He, Tian and Tao}]{dai2022dispfl}
\bibinfo{author}{Dai, R.}, \bibinfo{author}{Shen, L.}, \bibinfo{author}{He, F.}, \bibinfo{author}{Tian, X.}, \bibinfo{author}{Tao, D.}, \bibinfo{year}{2022}.
\newblock \bibinfo{title}{Dispfl: Towards communication-efficient personalized federated learning via decentralized sparse training}, in: \bibinfo{booktitle}{International conference on machine learning}, \bibinfo{organization}{PMLR}. pp. \bibinfo{pages}{4587--4604}.
\bibitem[{Fallah et~al.(2020)Fallah, Mokhtari and Ozdaglar}]{NEURIPS2020_24389bfe}
\bibinfo{author}{Fallah, A.}, \bibinfo{author}{Mokhtari, A.}, \bibinfo{author}{Ozdaglar, A.}, \bibinfo{year}{2020}.
\newblock \bibinfo{title}{Personalized federated learning with theoretical guarantees: A model-agnostic meta-learning approach}, in: \bibinfo{editor}{Larochelle, H.}, \bibinfo{editor}{Ranzato, M.}, \bibinfo{editor}{Hadsell, R.}, \bibinfo{editor}{Balcan, M.}, \bibinfo{editor}{Lin, H.} (Eds.), \bibinfo{booktitle}{Advances in Neural Information Processing Systems}, \bibinfo{publisher}{Curran Associates, Inc.}. pp. \bibinfo{pages}{3557--3568}.
\bibitem[{Ficco et~al.(2024)Ficco, Guerriero, Milite, Palmieri, Pietrantuono and Russo}]{FICCO2024102189}
\bibinfo{author}{Ficco, M.}, \bibinfo{author}{Guerriero, A.}, \bibinfo{author}{Milite, E.}, \bibinfo{author}{Palmieri, F.}, \bibinfo{author}{Pietrantuono, R.}, \bibinfo{author}{Russo, S.}, \bibinfo{year}{2024}.
\newblock \bibinfo{title}{Federated learning for iot devices: Enhancing tinyml with on-board training}.
\newblock \bibinfo{journal}{Information Fusion} \bibinfo{volume}{104}, \bibinfo{pages}{102189}.
\newblock \URLprefix \url{https://www.sciencedirect.com/science/article/pii/S1566253523005055}, \DOIprefix\doi{https://doi.org/10.1016/j.inffus.2023.102189}.
\bibitem[{Gal and Ghahramani(2015)}]{gal2015dropout}
\bibinfo{author}{Gal, Y.}, \bibinfo{author}{Ghahramani, Z.}, \bibinfo{year}{2015}.
\newblock \bibinfo{title}{Dropout as a bayesian approximation: Insights and applications}, in: \bibinfo{booktitle}{deep learning workshop, ICML}, p.~\bibinfo{pages}{2}.
\bibitem[{Gal and Ghahramani(2016)}]{pmlr-v48-gal16}
\bibinfo{author}{Gal, Y.}, \bibinfo{author}{Ghahramani, Z.}, \bibinfo{year}{2016}.
\newblock \bibinfo{title}{Dropout as a bayesian approximation: Representing model uncertainty in deep learning}, in: \bibinfo{editor}{Balcan, M.F.}, \bibinfo{editor}{Weinberger, K.Q.} (Eds.), \bibinfo{booktitle}{Proceedings of The 33rd International Conference on Machine Learning}, \bibinfo{publisher}{PMLR}, \bibinfo{address}{New York, New York, USA}. pp. \bibinfo{pages}{1050--1059}.
\newblock \URLprefix \url{https://proceedings.mlr.press/v48/gal16.html}.
\bibitem[{Gholami et~al.(2022)Gholami, Torkzaban and Baras}]{gholami2022trusted}
\bibinfo{author}{Gholami, A.}, \bibinfo{author}{Torkzaban, N.}, \bibinfo{author}{Baras, J.S.}, \bibinfo{year}{2022}.
\newblock \bibinfo{title}{Trusted decentralized federated learning}, in: \bibinfo{booktitle}{2022 IEEE 19th Annual Consumer Communications \& Networking Conference (CCNC)}, \bibinfo{organization}{IEEE}. pp. \bibinfo{pages}{1--6}.
\bibitem[{Ghosh et~al.(2022)Ghosh, Chung, Yin and Ramchandran}]{ifca}
\bibinfo{author}{Ghosh, A.}, \bibinfo{author}{Chung, J.}, \bibinfo{author}{Yin, D.}, \bibinfo{author}{Ramchandran, K.}, \bibinfo{year}{2022}.
\newblock \bibinfo{title}{An efficient framework for clustered federated learning}.
\newblock \bibinfo{journal}{IEEE Transactions on Information Theory} \bibinfo{volume}{68}, \bibinfo{pages}{8076--8091}.
\newblock \DOIprefix\doi{10.1109/TIT.2022.3192506}.
\bibitem[{Guo et~al.(2023)Guo, Qi, Qi, Wu and Li}]{GUO2023126831}
\bibinfo{author}{Guo, Q.}, \bibinfo{author}{Qi, Y.}, \bibinfo{author}{Qi, S.}, \bibinfo{author}{Wu, D.}, \bibinfo{author}{Li, Q.}, \bibinfo{year}{2023}.
\newblock \bibinfo{title}{Fedmcsa: Personalized federated learning via model components self-attention}.
\newblock \bibinfo{journal}{Neurocomputing} \bibinfo{volume}{560}, \bibinfo{pages}{126831}.
\newblock \URLprefix \url{https://www.sciencedirect.com/science/article/pii/S0925231223009542}, \DOIprefix\doi{https://doi.org/10.1016/j.neucom.2023.126831}.
\bibitem[{Hanzely et~al.(2022)Hanzely, Zhao et~al.}]{hanzely2022personalized}
\bibinfo{author}{Hanzely, F.}, \bibinfo{author}{Zhao, B.}, et~al., \bibinfo{year}{2022}.
\newblock \bibinfo{title}{Personalized federated learning: A unified framework and universal optimization techniques}.
\newblock \bibinfo{journal}{Transactions on Machine Learning Research} .
\bibitem[{He et~al.(2016)He, Zhang, Ren and Sun}]{he2016deep}
\bibinfo{author}{He, K.}, \bibinfo{author}{Zhang, X.}, \bibinfo{author}{Ren, S.}, \bibinfo{author}{Sun, J.}, \bibinfo{year}{2016}.
\newblock \bibinfo{title}{Deep residual learning for image recognition}, in: \bibinfo{booktitle}{Proceedings of the IEEE conference on computer vision and pattern recognition}, pp. \bibinfo{pages}{770--778}.
\bibitem[{Huang et~al.(2023)Huang, Luo, Liu, Zhao and Fu}]{HUANG2023118943}
\bibinfo{author}{Huang, X.}, \bibinfo{author}{Luo, Y.}, \bibinfo{author}{Liu, L.}, \bibinfo{author}{Zhao, W.}, \bibinfo{author}{Fu, S.}, \bibinfo{year}{2023}.
\newblock \bibinfo{title}{Randomization is all you need: A privacy-preserving federated learning framework for news recommendation}.
\newblock \bibinfo{journal}{Information Sciences} \bibinfo{volume}{637}, \bibinfo{pages}{118943}.
\newblock \URLprefix \url{https://www.sciencedirect.com/science/article/pii/S0020025523005121}, \DOIprefix\doi{https://doi.org/10.1016/j.ins.2023.118943}.
\bibitem[{Iman et~al.(2023)Iman, Arabnia and Rasheed}]{technologies11020040}
\bibinfo{author}{Iman, M.}, \bibinfo{author}{Arabnia, H.R.}, \bibinfo{author}{Rasheed, K.}, \bibinfo{year}{2023}.
\newblock \bibinfo{title}{A review of deep transfer learning and recent advancements}.
\newblock \bibinfo{journal}{Technologies} \bibinfo{volume}{11}.
\newblock \URLprefix \url{https://www.mdpi.com/2227-7080/11/2/40}, \DOIprefix\doi{10.3390/technologies11020040}.
\bibitem[{Imteaj et~al.(2022)Imteaj, Thakker, Wang, Li and Amini}]{9475501}
\bibinfo{author}{Imteaj, A.}, \bibinfo{author}{Thakker, U.}, \bibinfo{author}{Wang, S.}, \bibinfo{author}{Li, J.}, \bibinfo{author}{Amini, M.H.}, \bibinfo{year}{2022}.
\newblock \bibinfo{title}{A survey on federated learning for resource-constrained iot devices}.
\newblock \bibinfo{journal}{IEEE Internet of Things Journal} \bibinfo{volume}{9}, \bibinfo{pages}{1--24}.
\newblock \DOIprefix\doi{10.1109/JIOT.2021.3095077}.
\bibitem[{Jeong and Kountouris(2023)}]{10279714}
\bibinfo{author}{Jeong, E.}, \bibinfo{author}{Kountouris, M.}, \bibinfo{year}{2023}.
\newblock \bibinfo{title}{Personalized decentralized federated learning with knowledge distillation}, in: \bibinfo{booktitle}{ICC 2023 - IEEE International Conference on Communications}, pp. \bibinfo{pages}{1982--1987}.
\newblock \DOIprefix\doi{10.1109/ICC45041.2023.10279714}.
\bibitem[{Kalra et~al.(2023)Kalra, Wen, Cresswell, Volkovs and Tizhoosh}]{kalra2023decentralized}
\bibinfo{author}{Kalra, S.}, \bibinfo{author}{Wen, J.}, \bibinfo{author}{Cresswell, J.C.}, \bibinfo{author}{Volkovs, M.}, \bibinfo{author}{Tizhoosh, H.R.}, \bibinfo{year}{2023}.
\newblock \bibinfo{title}{Decentralized federated learning through proxy model sharing}.
\newblock \bibinfo{journal}{Nature communications} \bibinfo{volume}{14}, \bibinfo{pages}{2899}.
\bibitem[{Krizhevsky and Hinton(2009)}]{cifar10}
\bibinfo{author}{Krizhevsky, A.}, \bibinfo{author}{Hinton, G.}, \bibinfo{year}{2009}.
\newblock \bibinfo{title}{Learning multiple layers of features from tiny images}.
\newblock \bibinfo{journal}{Master's thesis, Department of Computer Science, University of Toronto} .
\bibitem[{Kullback and Leibler(1951)}]{kullback1951information}
\bibinfo{author}{Kullback, S.}, \bibinfo{author}{Leibler, R.A.}, \bibinfo{year}{1951}.
\newblock \bibinfo{title}{On information and sufficiency}.
\newblock \bibinfo{journal}{The annals of mathematical statistics} \bibinfo{volume}{22}, \bibinfo{pages}{79--86}.
\bibitem[{Lalitha et~al.(2018)Lalitha, Shekhar, Javidi and Koushanfar}]{lalitha2018fully}
\bibinfo{author}{Lalitha, A.}, \bibinfo{author}{Shekhar, S.}, \bibinfo{author}{Javidi, T.}, \bibinfo{author}{Koushanfar, F.}, \bibinfo{year}{2018}.
\newblock \bibinfo{title}{Fully decentralized federated learning}, in: \bibinfo{booktitle}{Third workshop on bayesian deep learning (NeurIPS)}.
\bibitem[{Li et~al.(2021a)Li, Li and Varshney}]{li2021decentralized}
\bibinfo{author}{Li, C.}, \bibinfo{author}{Li, G.}, \bibinfo{author}{Varshney, P.K.}, \bibinfo{year}{2021}a.
\newblock \bibinfo{title}{Decentralized federated learning via mutual knowledge transfer}.
\newblock \bibinfo{journal}{IEEE Internet of Things Journal} \bibinfo{volume}{9}, \bibinfo{pages}{1136--1147}.
\bibitem[{Li et~al.(2021b)Li, He and Song}]{moon}
\bibinfo{author}{Li, Q.}, \bibinfo{author}{He, B.}, \bibinfo{author}{Song, D.}, \bibinfo{year}{2021}b.
\newblock \bibinfo{title}{Model-contrastive federated learning}, in: \bibinfo{booktitle}{Proceedings of the IEEE/CVF conference on computer vision and pattern recognition}, pp. \bibinfo{pages}{10713--10722}.
\bibitem[{Li et~al.(2021c)Li, Hu, Beirami and Smith}]{ditto}
\bibinfo{author}{Li, T.}, \bibinfo{author}{Hu, S.}, \bibinfo{author}{Beirami, A.}, \bibinfo{author}{Smith, V.}, \bibinfo{year}{2021}c.
\newblock \bibinfo{title}{Ditto: Fair and robust federated learning through personalization}, in: \bibinfo{booktitle}{International Conference on Machine Learning}, \bibinfo{organization}{PMLR}. pp. \bibinfo{pages}{6357--6368}.
\bibitem[{Liang et~al.(2020)Liang, Liu, Ziyin, Allen, Auerbach, Brent, Salakhutdinov and Morency}]{liang2020think}
\bibinfo{author}{Liang, P.P.}, \bibinfo{author}{Liu, T.}, \bibinfo{author}{Ziyin, L.}, \bibinfo{author}{Allen, N.B.}, \bibinfo{author}{Auerbach, R.P.}, \bibinfo{author}{Brent, D.}, \bibinfo{author}{Salakhutdinov, R.}, \bibinfo{author}{Morency, L.P.}, \bibinfo{year}{2020}.
\newblock \bibinfo{title}{Think locally, act globally: Federated learning with local and global representations}.
\newblock \bibinfo{journal}{arXiv preprint arXiv:2001.01523} .
\bibitem[{Liu et~al.(2024)Liu, Li, Li and Sun}]{LIU2024}
\bibinfo{author}{Liu, H.}, \bibinfo{author}{Li, S.}, \bibinfo{author}{Li, W.}, \bibinfo{author}{Sun, W.}, \bibinfo{year}{2024}.
\newblock \bibinfo{title}{Efficient decentralized optimization for edge-enabled smart manufacturing: A federated learning-based framework}.
\newblock \bibinfo{journal}{Future Generation Computer Systems} \URLprefix \url{https://www.sciencedirect.com/science/article/pii/S0167739X24001146}, \DOIprefix\doi{https://doi.org/10.1016/j.future.2024.03.043}.
\bibitem[{Liu et~al.(2020)Liu, Ai, Sun, Zhang, Liu and Yu}]{liu2020fedcoin}
\bibinfo{author}{Liu, Y.}, \bibinfo{author}{Ai, Z.}, \bibinfo{author}{Sun, S.}, \bibinfo{author}{Zhang, S.}, \bibinfo{author}{Liu, Z.}, \bibinfo{author}{Yu, H.}, \bibinfo{year}{2020}.
\newblock \bibinfo{title}{Fedcoin: A peer-to-peer payment system for federated learning}, in: \bibinfo{booktitle}{Federated learning: privacy and incentive}. \bibinfo{publisher}{Springer}, pp. \bibinfo{pages}{125--138}.
\bibitem[{Long et~al.(2023)Long, Xie, Shen, Zhou, Wang and Jiang}]{long2023multi}
\bibinfo{author}{Long, G.}, \bibinfo{author}{Xie, M.}, \bibinfo{author}{Shen, T.}, \bibinfo{author}{Zhou, T.}, \bibinfo{author}{Wang, X.}, \bibinfo{author}{Jiang, J.}, \bibinfo{year}{2023}.
\newblock \bibinfo{title}{Multi-center federated learning: clients clustering for better personalization}.
\newblock \bibinfo{journal}{World Wide Web} \bibinfo{volume}{26}, \bibinfo{pages}{481--500}.
\bibitem[{Ma et~al.(2022)Ma, Xu, Xu, Liu and Xue}]{ma2022like}
\bibinfo{author}{Ma, Z.}, \bibinfo{author}{Xu, Y.}, \bibinfo{author}{Xu, H.}, \bibinfo{author}{Liu, J.}, \bibinfo{author}{Xue, Y.}, \bibinfo{year}{2022}.
\newblock \bibinfo{title}{Like attracts like: Personalized federated learning in decentralized edge computing}.
\newblock \bibinfo{journal}{IEEE Transactions on Mobile Computing} .
\bibitem[{Martínez~Beltrán et~al.(2023)Martínez~Beltrán, Pérez, Sánchez, Bernal, Bovet, Pérez, Pérez and Celdrán}]{10251949}
\bibinfo{author}{Martínez~Beltrán, E.T.}, \bibinfo{author}{Pérez, M.Q.}, \bibinfo{author}{Sánchez, P.M.S.}, \bibinfo{author}{Bernal, S.L.}, \bibinfo{author}{Bovet, G.}, \bibinfo{author}{Pérez, M.G.}, \bibinfo{author}{Pérez, G.M.}, \bibinfo{author}{Celdrán, A.H.}, \bibinfo{year}{2023}.
\newblock \bibinfo{title}{Decentralized federated learning: Fundamentals, state of the art, frameworks, trends, and challenges}.
\newblock \bibinfo{journal}{IEEE Communications Surveys \& Tutorials} \bibinfo{volume}{25}, \bibinfo{pages}{2983--3013}.
\newblock \DOIprefix\doi{10.1109/COMST.2023.3315746}.
\bibitem[{McMahan et~al.(2017)McMahan, Moore, Ramage, Hampson and Arcas}]{fedavg}
\bibinfo{author}{McMahan, B.}, \bibinfo{author}{Moore, E.}, \bibinfo{author}{Ramage, D.}, \bibinfo{author}{Hampson, S.}, \bibinfo{author}{Arcas, B.A.y.}, \bibinfo{year}{2017}.
\newblock \bibinfo{title}{Communication-efficient learning of deep networks from decentralized data}, in: \bibinfo{editor}{Singh, A.}, \bibinfo{editor}{Zhu, J.} (Eds.), \bibinfo{booktitle}{Proceedings of the 20th International Conference on Artificial Intelligence and Statistics}, \bibinfo{publisher}{PMLR}. pp. \bibinfo{pages}{1273--1282}.
\newblock \URLprefix \url{https://proceedings.mlr.press/v54/mcmahan17a.html}.
\bibitem[{Men{\'e}ndez et~al.(1997)Men{\'e}ndez, Pardo, Pardo and Pardo}]{menendez1997jensen}
\bibinfo{author}{Men{\'e}ndez, M.}, \bibinfo{author}{Pardo, J.}, \bibinfo{author}{Pardo, L.}, \bibinfo{author}{Pardo, M.}, \bibinfo{year}{1997}.
\newblock \bibinfo{title}{The jensen-shannon divergence}.
\newblock \bibinfo{journal}{Journal of the Franklin Institute} \bibinfo{volume}{334}, \bibinfo{pages}{307--318}.
\bibitem[{Morafah et~al.(2023a)Morafah, Vahidian, Wang and Lin}]{10081485}
\bibinfo{author}{Morafah, M.}, \bibinfo{author}{Vahidian, S.}, \bibinfo{author}{Wang, W.}, \bibinfo{author}{Lin, B.}, \bibinfo{year}{2023}a.
\newblock \bibinfo{title}{Flis: Clustered federated learning via inference similarity for non-iid data distribution}.
\newblock \bibinfo{journal}{IEEE Open Journal of the Computer Society} \bibinfo{volume}{4}, \bibinfo{pages}{109--120}.
\newblock \DOIprefix\doi{10.1109/OJCS.2023.3262203}.
\bibitem[{Morafah et~al.(2023b)Morafah, Vahidian, Wang and Lin}]{flis}
\bibinfo{author}{Morafah, M.}, \bibinfo{author}{Vahidian, S.}, \bibinfo{author}{Wang, W.}, \bibinfo{author}{Lin, B.}, \bibinfo{year}{2023}b.
\newblock \bibinfo{title}{Flis: Clustered federated learning via inference similarity for non-iid data distribution}.
\newblock \bibinfo{journal}{IEEE Open Journal of the Computer Society} \bibinfo{volume}{4}, \bibinfo{pages}{109--120}.
\bibitem[{Na et~al.(2024)Na, Liang and Yiu}]{NA2024127630}
\bibinfo{author}{Na, S.}, \bibinfo{author}{Liang, Y.}, \bibinfo{author}{Yiu, S.M.}, \bibinfo{year}{2024}.
\newblock \bibinfo{title}{A federated learning incentive mechanism in a non-monopoly market}.
\newblock \bibinfo{journal}{Neurocomputing} \bibinfo{volume}{586}, \bibinfo{pages}{127630}.
\newblock \URLprefix \url{https://www.sciencedirect.com/science/article/pii/S0925231224004016}, \DOIprefix\doi{https://doi.org/10.1016/j.neucom.2024.127630}.
\bibitem[{Netzer et~al.(2011)Netzer, Wang, Coates, Bissacco, Wu, Ng et~al.}]{svhn}
\bibinfo{author}{Netzer, Y.}, \bibinfo{author}{Wang, T.}, \bibinfo{author}{Coates, A.}, \bibinfo{author}{Bissacco, A.}, \bibinfo{author}{Wu, B.}, \bibinfo{author}{Ng, A.Y.}, et~al., \bibinfo{year}{2011}.
\newblock \bibinfo{title}{Reading digits in natural images with unsupervised feature learning}, in: \bibinfo{booktitle}{NIPS workshop on deep learning and unsupervised feature learning}, \bibinfo{organization}{Granada, Spain}. p.~\bibinfo{pages}{7}.
\bibitem[{Pillutla et~al.(2022)Pillutla, Malik, Mohamed, Rabbat, Sanjabi and Xiao}]{fedalt}
\bibinfo{author}{Pillutla, K.}, \bibinfo{author}{Malik, K.}, \bibinfo{author}{Mohamed, A.R.}, \bibinfo{author}{Rabbat, M.}, \bibinfo{author}{Sanjabi, M.}, \bibinfo{author}{Xiao, L.}, \bibinfo{year}{2022}.
\newblock \bibinfo{title}{Federated learning with partial model personalization}, in: \bibinfo{editor}{Chaudhuri, K.}, \bibinfo{editor}{Jegelka, S.}, \bibinfo{editor}{Song, L.}, \bibinfo{editor}{Szepesvari, C.}, \bibinfo{editor}{Niu, G.}, \bibinfo{editor}{Sabato, S.} (Eds.), \bibinfo{booktitle}{Proceedings of the 39th International Conference on Machine Learning}, \bibinfo{publisher}{PMLR}. pp. \bibinfo{pages}{17716--17758}.
\newblock \URLprefix \url{https://proceedings.mlr.press/v162/pillutla22a.html}.
\bibitem[{Qu et~al.(2021)Qu, Dai, Zhuang, Chen, Dong, Wu and Guo}]{9687521}
\bibinfo{author}{Qu, Y.}, \bibinfo{author}{Dai, H.}, \bibinfo{author}{Zhuang, Y.}, \bibinfo{author}{Chen, J.}, \bibinfo{author}{Dong, C.}, \bibinfo{author}{Wu, F.}, \bibinfo{author}{Guo, S.}, \bibinfo{year}{2021}.
\newblock \bibinfo{title}{Decentralized federated learning for uav networks: Architecture, challenges, and opportunities}.
\newblock \bibinfo{journal}{IEEE Network} \bibinfo{volume}{35}, \bibinfo{pages}{156--162}.
\newblock \DOIprefix\doi{10.1109/MNET.001.2100253}.
\bibitem[{Roy et~al.(2019)Roy, Siddiqui, P{\"o}lsterl, Navab and Wachinger}]{roy2019braintorrent}
\bibinfo{author}{Roy, A.G.}, \bibinfo{author}{Siddiqui, S.}, \bibinfo{author}{P{\"o}lsterl, S.}, \bibinfo{author}{Navab, N.}, \bibinfo{author}{Wachinger, C.}, \bibinfo{year}{2019}.
\newblock \bibinfo{title}{Braintorrent: A peer-to-peer environment for decentralized federated learning}.
\newblock \bibinfo{journal}{arXiv preprint arXiv:1905.06731} .
\bibitem[{Sattler et~al.(2020)Sattler, M{\"u}ller and Samek}]{cfl}
\bibinfo{author}{Sattler, F.}, \bibinfo{author}{M{\"u}ller, K.R.}, \bibinfo{author}{Samek, W.}, \bibinfo{year}{2020}.
\newblock \bibinfo{title}{Clustered federated learning: Model-agnostic distributed multitask optimization under privacy constraints}.
\newblock \bibinfo{journal}{IEEE transactions on neural networks and learning systems} \bibinfo{volume}{32}, \bibinfo{pages}{3710--3722}.
\bibitem[{Shen et~al.(2023)Shen, Fu, Gui, Susilo and Zhang}]{SHEN2023119261}
\bibinfo{author}{Shen, G.}, \bibinfo{author}{Fu, Z.}, \bibinfo{author}{Gui, Y.}, \bibinfo{author}{Susilo, W.}, \bibinfo{author}{Zhang, M.}, \bibinfo{year}{2023}.
\newblock \bibinfo{title}{Efficient and privacy-preserving online diagnosis scheme based on federated learning in e-healthcare system}.
\newblock \bibinfo{journal}{Information Sciences} \bibinfo{volume}{647}, \bibinfo{pages}{119261}.
\newblock \URLprefix \url{https://www.sciencedirect.com/science/article/pii/S0020025523008460}, \DOIprefix\doi{https://doi.org/10.1016/j.ins.2023.119261}.
\bibitem[{Simonyan and Zisserman(2014)}]{simonyan2014very}
\bibinfo{author}{Simonyan, K.}, \bibinfo{author}{Zisserman, A.}, \bibinfo{year}{2014}.
\newblock \bibinfo{title}{Very deep convolutional networks for large-scale image recognition}.
\newblock \bibinfo{journal}{arXiv preprint arXiv:1409.1556} .
\bibitem[{Srivastava et~al.(2014)Srivastava, Hinton, Krizhevsky, Sutskever and Salakhutdinov}]{dropout}
\bibinfo{author}{Srivastava, N.}, \bibinfo{author}{Hinton, G.}, \bibinfo{author}{Krizhevsky, A.}, \bibinfo{author}{Sutskever, I.}, \bibinfo{author}{Salakhutdinov, R.}, \bibinfo{year}{2014}.
\newblock \bibinfo{title}{Dropout: a simple way to prevent neural networks from overfitting}.
\newblock \bibinfo{journal}{The journal of machine learning research} \bibinfo{volume}{15}, \bibinfo{pages}{1929--1958}.
\bibitem[{Sun et~al.(2023)Sun, Li and Wang}]{9850408}
\bibinfo{author}{Sun, T.}, \bibinfo{author}{Li, D.}, \bibinfo{author}{Wang, B.}, \bibinfo{year}{2023}.
\newblock \bibinfo{title}{Decentralized federated averaging}.
\newblock \bibinfo{journal}{IEEE Transactions on Pattern Analysis and Machine Intelligence} \bibinfo{volume}{45}, \bibinfo{pages}{4289--4301}.
\newblock \DOIprefix\doi{10.1109/TPAMI.2022.3196503}.
\bibitem[{Tang et~al.(2022)Tang, Shi, Li and Chu}]{tang2022gossipfl}
\bibinfo{author}{Tang, Z.}, \bibinfo{author}{Shi, S.}, \bibinfo{author}{Li, B.}, \bibinfo{author}{Chu, X.}, \bibinfo{year}{2022}.
\newblock \bibinfo{title}{Gossipfl: A decentralized federated learning framework with sparsified and adaptive communication}.
\newblock \bibinfo{journal}{IEEE Transactions on Parallel and Distributed Systems} \bibinfo{volume}{34}, \bibinfo{pages}{909--922}.
\bibitem[{Wang et~al.(2024)Wang, Yang, Cui, Che, Lyu, Xu and Ma}]{wang2024towards}
\bibinfo{author}{Wang, J.}, \bibinfo{author}{Yang, X.}, \bibinfo{author}{Cui, S.}, \bibinfo{author}{Che, L.}, \bibinfo{author}{Lyu, L.}, \bibinfo{author}{Xu, D.D.}, \bibinfo{author}{Ma, F.}, \bibinfo{year}{2024}.
\newblock \bibinfo{title}{Towards personalized federated learning via heterogeneous model reassembly}.
\newblock \bibinfo{journal}{Advances in Neural Information Processing Systems} \bibinfo{volume}{36}.
\bibitem[{Wang et~al.(2022)Wang, Cheng, Luo, Yu, Ni, Tong, Chen and Zhang}]{10027784}
\bibinfo{author}{Wang, T.}, \bibinfo{author}{Cheng, W.}, \bibinfo{author}{Luo, D.}, \bibinfo{author}{Yu, W.}, \bibinfo{author}{Ni, J.}, \bibinfo{author}{Tong, L.}, \bibinfo{author}{Chen, H.}, \bibinfo{author}{Zhang, X.}, \bibinfo{year}{2022}.
\newblock \bibinfo{title}{Personalized federated learning via heterogeneous modular networks}, in: \bibinfo{booktitle}{2022 IEEE International Conference on Data Mining (ICDM)}, pp. \bibinfo{pages}{1197--1202}.
\newblock \DOIprefix\doi{10.1109/ICDM54844.2022.00154}.
\bibitem[{Weng et~al.(2019)Weng, Weng, Zhang, Li, Zhang and Luo}]{weng2019deepchain}
\bibinfo{author}{Weng, J.}, \bibinfo{author}{Weng, J.}, \bibinfo{author}{Zhang, J.}, \bibinfo{author}{Li, M.}, \bibinfo{author}{Zhang, Y.}, \bibinfo{author}{Luo, W.}, \bibinfo{year}{2019}.
\newblock \bibinfo{title}{Deepchain: Auditable and privacy-preserving deep learning with blockchain-based incentive}.
\newblock \bibinfo{journal}{IEEE Transactions on Dependable and Secure Computing} \bibinfo{volume}{18}, \bibinfo{pages}{2438--2455}.
\bibitem[{Witt et~al.(2023)Witt, Heyer, Toyoda, Samek and Li}]{9997105}
\bibinfo{author}{Witt, L.}, \bibinfo{author}{Heyer, M.}, \bibinfo{author}{Toyoda, K.}, \bibinfo{author}{Samek, W.}, \bibinfo{author}{Li, D.}, \bibinfo{year}{2023}.
\newblock \bibinfo{title}{Decentral and incentivized federated learning frameworks: A systematic literature review}.
\newblock \bibinfo{journal}{IEEE Internet of Things Journal} \bibinfo{volume}{10}, \bibinfo{pages}{3642--3663}.
\newblock \DOIprefix\doi{10.1109/JIOT.2022.3231363}.
\bibitem[{Wu et~al.(2022)Wu, Wu, Lyu, Huang and Xie}]{wu2022communication}
\bibinfo{author}{Wu, C.}, \bibinfo{author}{Wu, F.}, \bibinfo{author}{Lyu, L.}, \bibinfo{author}{Huang, Y.}, \bibinfo{author}{Xie, X.}, \bibinfo{year}{2022}.
\newblock \bibinfo{title}{Communication-efficient federated learning via knowledge distillation}.
\newblock \bibinfo{journal}{Nature communications} \bibinfo{volume}{13}, \bibinfo{pages}{2032}.
\bibitem[{Xu and Fan(2023)}]{10182241}
\bibinfo{author}{Xu, Y.}, \bibinfo{author}{Fan, H.}, \bibinfo{year}{2023}.
\newblock \bibinfo{title}{Feddk: Improving cyclic knowledge distillation for personalized healthcare federated learning}.
\newblock \bibinfo{journal}{IEEE Access} \bibinfo{volume}{11}, \bibinfo{pages}{72409--72417}.
\newblock \DOIprefix\doi{10.1109/ACCESS.2023.3294812}.
\bibitem[{Yan et~al.(2023)Yan, Wang, Yuan and Li}]{yan2023criticalfl}
\bibinfo{author}{Yan, G.}, \bibinfo{author}{Wang, H.}, \bibinfo{author}{Yuan, X.}, \bibinfo{author}{Li, J.}, \bibinfo{year}{2023}.
\newblock \bibinfo{title}{Criticalfl: A critical learning periods augmented client selection framework for efficient federated learning}, in: \bibinfo{booktitle}{Proceedings of the 29th ACM SIGKDD Conference on Knowledge Discovery and Data Mining}, pp. \bibinfo{pages}{2898--2907}.
\bibitem[{Yang et~al.(2023)Yang, Shi, Wei, Liu, Zhao, Ke, Pfister and Ni}]{medmnistv2}
\bibinfo{author}{Yang, J.}, \bibinfo{author}{Shi, R.}, \bibinfo{author}{Wei, D.}, \bibinfo{author}{Liu, Z.}, \bibinfo{author}{Zhao, L.}, \bibinfo{author}{Ke, B.}, \bibinfo{author}{Pfister, H.}, \bibinfo{author}{Ni, B.}, \bibinfo{year}{2023}.
\newblock \bibinfo{title}{Medmnist v2-a large-scale lightweight benchmark for 2d and 3d biomedical image classification}.
\newblock \bibinfo{journal}{Scientific Data} \bibinfo{volume}{10}, \bibinfo{pages}{41}.
\bibitem[{Ye et~al.(2023)Ye, Wei, Cui, Chen, Fu and Gao}]{rcfl}
\bibinfo{author}{Ye, T.}, \bibinfo{author}{Wei, S.}, \bibinfo{author}{Cui, J.}, \bibinfo{author}{Chen, C.}, \bibinfo{author}{Fu, Y.}, \bibinfo{author}{Gao, M.}, \bibinfo{year}{2023}.
\newblock \bibinfo{title}{Robust clustered federated learning}, in: \bibinfo{booktitle}{International Conference on Database Systems for Advanced Applications}, \bibinfo{organization}{Springer}. pp. \bibinfo{pages}{677--692}.
\bibitem[{Yu et~al.(2024)Yu, Wang, Zeng, Zhao and Yu}]{YU2024127290}
\bibinfo{author}{Yu, F.}, \bibinfo{author}{Wang, L.}, \bibinfo{author}{Zeng, B.}, \bibinfo{author}{Zhao, K.}, \bibinfo{author}{Yu, R.}, \bibinfo{year}{2024}.
\newblock \bibinfo{title}{Personalized and privacy-enhanced federated learning framework via knowledge distillation}.
\newblock \bibinfo{journal}{Neurocomputing} \bibinfo{volume}{575}, \bibinfo{pages}{127290}.
\newblock \URLprefix \url{https://www.sciencedirect.com/science/article/pii/S0925231224000614}, \DOIprefix\doi{https://doi.org/10.1016/j.neucom.2024.127290}.
\bibitem[{Zhang et~al.(2021)Zhang, Bengio, Hardt, Recht and Vinyals}]{nn_has_memory}
\bibinfo{author}{Zhang, C.}, \bibinfo{author}{Bengio, S.}, \bibinfo{author}{Hardt, M.}, \bibinfo{author}{Recht, B.}, \bibinfo{author}{Vinyals, O.}, \bibinfo{year}{2021}.
\newblock \bibinfo{title}{Understanding deep learning (still) requires rethinking generalization}.
\newblock \bibinfo{journal}{Commun. ACM} \bibinfo{volume}{64}, \bibinfo{pages}{107–115}.
\newblock \URLprefix \url{https://doi.org/10.1145/3446776}, \DOIprefix\doi{10.1145/3446776}.
\bibitem[{Zhang et~al.(2023)Zhang, Meng, Liu, Wu, Wang and Ning}]{ZHANG2023126791}
\bibinfo{author}{Zhang, C.}, \bibinfo{author}{Meng, X.}, \bibinfo{author}{Liu, Q.}, \bibinfo{author}{Wu, S.}, \bibinfo{author}{Wang, L.}, \bibinfo{author}{Ning, H.}, \bibinfo{year}{2023}.
\newblock \bibinfo{title}{Fedbrain: A robust multi-site brain network analysis framework based on federated learning for brain disease diagnosis}.
\newblock \bibinfo{journal}{Neurocomputing} \bibinfo{volume}{559}, \bibinfo{pages}{126791}.
\newblock \URLprefix \url{https://www.sciencedirect.com/science/article/pii/S0925231223009141}, \DOIprefix\doi{https://doi.org/10.1016/j.neucom.2023.126791}.
\bibitem[{Zhang et~al.(2022)Zhang, Li, Li, Guo and Shao}]{zhang2022personalized}
\bibinfo{author}{Zhang, X.}, \bibinfo{author}{Li, Y.}, \bibinfo{author}{Li, W.}, \bibinfo{author}{Guo, K.}, \bibinfo{author}{Shao, Y.}, \bibinfo{year}{2022}.
\newblock \bibinfo{title}{Personalized federated learning via variational bayesian inference}, in: \bibinfo{booktitle}{International Conference on Machine Learning}, \bibinfo{organization}{PMLR}. pp. \bibinfo{pages}{26293--26310}.
\bibitem[{Zheng et~al.(2018)Zheng, Xie, Dai, Chen and Wang}]{zheng2018blockchain}
\bibinfo{author}{Zheng, Z.}, \bibinfo{author}{Xie, S.}, \bibinfo{author}{Dai, H.N.}, \bibinfo{author}{Chen, X.}, \bibinfo{author}{Wang, H.}, \bibinfo{year}{2018}.
\newblock \bibinfo{title}{Blockchain challenges and opportunities: A survey}.
\newblock \bibinfo{journal}{International journal of web and grid services} \bibinfo{volume}{14}, \bibinfo{pages}{352--375}.
\bibitem[{Zhou et~al.(2023)Zhou, Huang, Li, Liu and Zheng}]{zhou2023comavg}
\bibinfo{author}{Zhou, S.}, \bibinfo{author}{Huang, H.}, \bibinfo{author}{Li, R.}, \bibinfo{author}{Liu, J.}, \bibinfo{author}{Zheng, Z.}, \bibinfo{year}{2023}.
\newblock \bibinfo{title}{Comavg: Robust decentralized federated learning with random committees}.
\newblock \bibinfo{journal}{Computer Communications} \bibinfo{volume}{211}, \bibinfo{pages}{147--156}.
\bibitem[{Zhu et~al.(2024)Zhu, Fan and Xie}]{fedtsdp}
\bibinfo{author}{Zhu, H.}, \bibinfo{author}{Fan, Y.}, \bibinfo{author}{Xie, Z.}, \bibinfo{year}{2024}.
\newblock \bibinfo{title}{Federated two-stage decoupling with adaptive personalization layers}.
\newblock \bibinfo{journal}{Complex \& Intelligent Systems} \URLprefix \url{https://doi.org/10.1007/s40747-024-01342-1}, \DOIprefix\doi{10.1007/s40747-024-01342-1}.
\bibitem[{Zhu et~al.(2021)Zhu, Xu, Liu and Jin}]{ZHU2021371}
\bibinfo{author}{Zhu, H.}, \bibinfo{author}{Xu, J.}, \bibinfo{author}{Liu, S.}, \bibinfo{author}{Jin, Y.}, \bibinfo{year}{2021}.
\newblock \bibinfo{title}{Federated learning on non-iid data: A survey}.
\newblock \bibinfo{journal}{Neurocomputing} \bibinfo{volume}{465}, \bibinfo{pages}{371--390}.
\newblock \URLprefix \url{https://www.sciencedirect.com/science/article/pii/S0925231221013254}, \DOIprefix\doi{https://doi.org/10.1016/j.neucom.2021.07.098}.

\end{thebibliography}
\end{document}